\documentclass{article}

\usepackage{arxiv}

\usepackage[utf8]{inputenc} 
\usepackage[T1]{fontenc}    
\usepackage{hyperref}       
\usepackage{url}            
\usepackage{booktabs}       
\usepackage{amsfonts}       
\usepackage{nicefrac}       
\usepackage{graphicx}
\usepackage{natbib}
\usepackage{doi}

\usepackage{float}
\usepackage{soul}
\usepackage{color}
\usepackage{tabularx} 
\usepackage{xcolor}
\usepackage{caption}
\usepackage{adjustbox}
\usepackage{amsmath}
\usepackage{amssymb}
\usepackage{tikz}
\usepackage{circuitikz}
\usepackage{pgfplots}
\usepackage{upgreek}
\usepackage{soul}
\pgfplotsset{compat=1.17}

\usepackage{pgfplots}
\usepackage{changepage}
\usepackage{multicol}
\usepackage{wrapfig}
\usepgfplotslibrary{fillbetween}
\usepackage{scontents}

\usepackage{lineno}
\usepackage{colortbl}
\usepackage{rotating}
\usepackage{pgfplots,pgfplotstable}
\usepackage{scalefnt}
\usepackage{subcaption}

\usepackage{setspace} 

\newtheorem{definition}{Definition}

\title{Collaborative Trustworthiness for Good Decision Making in Autonomous Systems}

\author{
    Selma Saidi \\
    Institute of Computer and Network Engineering (IDA) \\
    Technische Universität Braunschweig \\
    Braunschweig, Germany \\
    \texttt{saidi@ida.ing.tu-bs.de} \\
    \And
    Omar Laimona \\
    Institute of Computer and Network Engineering (IDA) \\
    Technische Universität Braunschweig \\
    Braunschweig, Germany \\
    \texttt{laimona@ida.ing.tu-bs.de} \\
    \And
    Christoph Schmickler \\
    Chair of Embedded Systems \\
    TU Dortmund University \\
    Dortmund, Germany \\
    \texttt{christoph.schmickler@tu-dortmund.de} \\
    \And
    Dirk Ziegenbein \\
    Robert Bosch Corporate Research \\
    Renningen, Germany \\
    \texttt{dirk.ziegenbein@de.bosch.com}
}

\date{}

\renewcommand{\headeright}{}
\renewcommand{\undertitle}{}

\hypersetup{
  pdftitle={Collaborative Trustworthiness for Autonomous Systems},
  pdfsubject={Artificial Intelligence, Autonomous Systems},
  pdfauthor={},
  pdfkeywords={Autonomous systems, Trustworthiness, BDD, Aggregation}
}

\begin{document}
\maketitle

\begin{abstract}
Autonomous systems are becoming an integral part of many application domains, like in the mobility sector.  
However, ensuring their safe and correct behaviour in dynamic
and complex environments remains a significant challenge, where systems should autonomously make decisions 
e.g., about manoeuvring. We propose in this paper a general collaborative approach for increasing the level of trustworthiness in the environment of operation and improve reliability and good decision making in autonomous system. 
In the presence of conflicting information, \emph{aggregation} becomes a major issue for trustworthy decision making based on collaborative data sharing. 
%
%
Unlike classical approaches in the literature that rely on consensus or majority as aggregation rule, we exploit the fact that autonomous systems have different quality attributes like perception quality. We use this criteria to determine which autonomous systems are trustworthy and borrow concepts from social epistemology to define aggregation and propagation rules, used for automated decision making. We use Binary Decision Diagrams (BDDs) as formal models for beliefs aggregation and propagation, and formulate reduction rules to reduce the size of the BDDs and allow efficient computation structures for collaborative automated reasoning. 



\end{abstract}

\section{Assuring Safe Autonomy Through Collaboration}
\label{sec:intro}


Autonomous systems have the ability to interact with their environment and act independently by solving complex tasks without human intervention. 
The main challenge with designing autonomous systems is to provide them with the ability to operate correctly and safely in dynamic and open contexts. That is, performing tasks at operation time in an environment that is uncertain or was not fully known or defined at design time. In such environments, autonomous systems must \emph{self-decide} e.g., about manoeuvring in self-driving cars.
When extended to fields such as automated driving, where safety is a critical requirement, assuring high levels of trustworthiness and "good" 
decision making in complex environments becomes a real challenge. In this paper, we consider collaboration between autonomous systems to increase trustworthiness and improve decision making. For that, we use concepts of beliefs aggregation and propagation from social sciences and social epistemology, where we exploit autonomous systems individual quality of attributes as a measure and evidence for trustworthiness during aggregation. We further define reliability as a quantitative measure of good decision making (i.e., less errors in perceiving the environment).  In the following, we first motivate and introduce the ideas and concepts used in this paper through an illustrative example, before we provide formal definitions and contribution in the subsequent section(s).





\begin{figure}
    \centering
    \includegraphics[scale = 0.75]{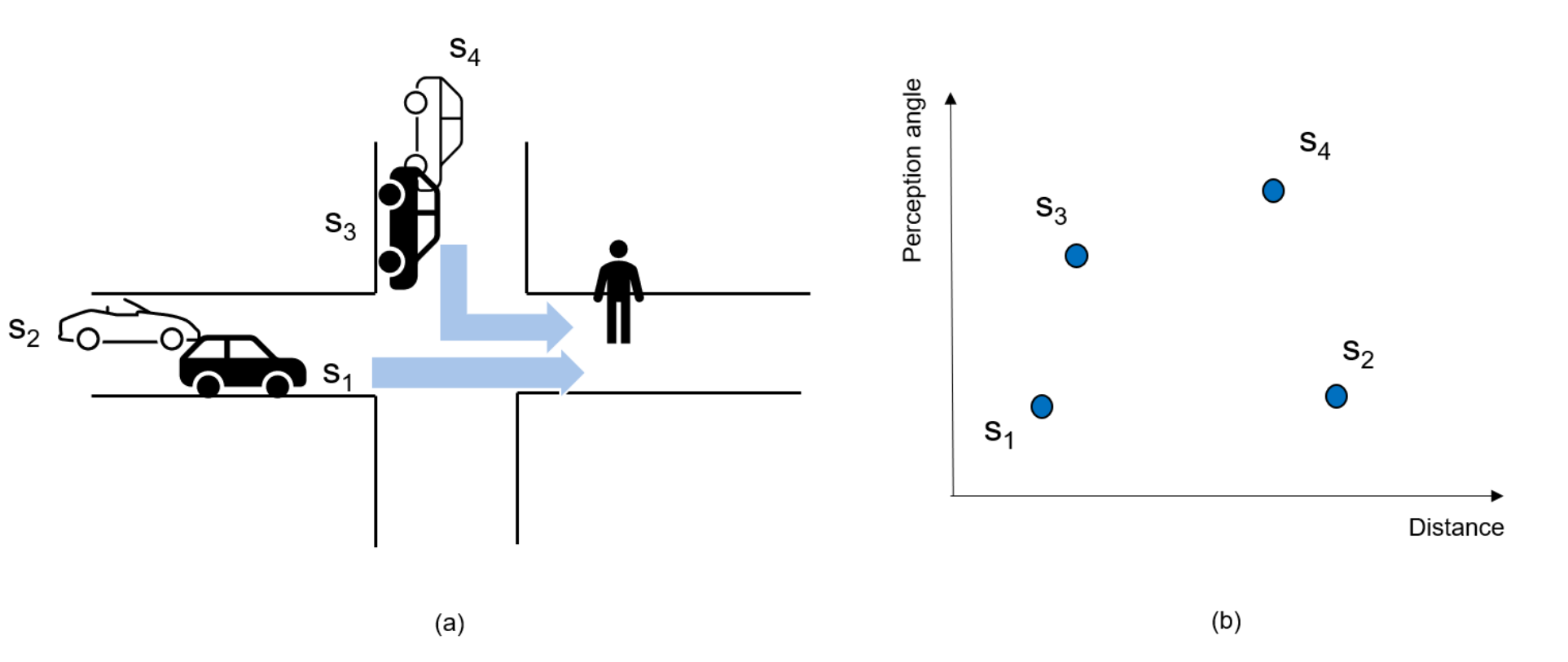} 
    \caption{
        Example of autonomous systems operating in the same environment a) vehicles with different quality attributes 
        (distance and perception angle) 
        possibly detecting a pedestrian, b) ranking between vehicles based on the quality of their attributes
        (assuming the smaller is the better).
   Vehicle $s_1$ (resp., $s_4$) has better (resp., worse) distance to the pedestrian and a better (resp., worse) perception angle compared to other vehicles. Both $s_2$ and $s_3$ are not comparable since $s_2$ has a better perception angle but a larger distance to the pedestrian compared to $s_3$, while for $s_3$ it is the opposite.
    }
    \label{fig:example}
\end{figure}

Let us consider the following simple example depicted in Fig.~\ref{fig:example}, of four autonomous vehicles operating in the same environment, such as a smart intersection, which contains a pedestrian. 
Every vehicle should make autonomous decisions 
about manoeuvring to reach its goal 
without colliding with neither the pedestrian nor the other vehicles. 
Assume the environment can be described using simple predicates of the form "a pedestrian has been detected". A more detailed predicate of the form "road segment X is blocked by pedestrian" 
 that includes a location for driving can as well be defined. Every autonomous vehicle can formulate, based on its perception systems, "True" or "False" statements, that we refer to as "beliefs\footnote{We use in this paper concepts from social epistemology and borrow some terminology 
 that we formally define in Sec.~\ref{subsec:def}.}",  to such predicates. 
When beliefs are \emph{collectively} shared
, they can contribute to a better decision making.
 Autonomous vehicles can use shared information to correct their beliefs e.g., resulting from perception errors, to change their assignment from a "False" to a "True" statement for a given predicate.
The main question becomes during aggregation which vehicle is trustworthy and its belief can be propagated and adopted in the presence of conflicting beliefs shared by other autonomous vehicles? We later show in Sec~\ref{sec:exp} that majority rule or consensus, predominantly adopted in classical autonomous multi-agents~\cite{multiagent}, do not necessarily lead to increased reliability, and moreover comes at the costs of a large communication overhead. We present in this paper an alternative approach where we exploit the different quality of attributes of autonomous systems as a measure and evidence for trustworthiness. 
For instance, every individual vehicle in the 
 example depicted in Fig.~\ref{fig:example}
 is at a different physical position in the intersection, 
therefore having a different distance and perception angle with respect to the pedestrian. 
%
Vehicle $s_1$ (resp., $s_4$) has better (resp., worse) distance to the pedestrian and a better (resp., worse) perception angle compared to other vehicles and can be considered as more (resp., less) trustworthy. 
we establish a ranking among autonomous systems based on their quality of attributes and therefore their level of trustworthiness. Based on that, we define \emph{rules} for beliefs aggregation and propagation to reach higher reliability and efficiency. 



The term \emph{collaborative autonomy} as defined by~\cite{Gill11} already recognizes that autonomous systems have different capabilities, for instance of pattern recognition, 
and maintain their roles and priorities as they apply their unique skills and leadership autonomy in a problem-solving process.
A consensus in reaching collaborative decisions 
(as traditionally applied in autonomous multi-agents) becomes therefore not always required as consensus usually assumes (implicitly or not) that autonomous agents are identical.
We exploit the idea that autonomous systems can be characterized using different quality attributes (e.g., sensors with different qualities, distance and perception angles) that can be used to establish a ranking among autonomous systems. 
The ranking is used to define autonomous systems that are more \emph{trustworthy} than others based on better quality of their attributes. 
Therefore, instead of broadcasting all beliefs and relying on the majority considering the entire group, we rely on a sub-group of most trustworthy autonomous systems 
to provide an aggregate function. The aggregated belief from the sub-group 
can later further propagate to the rest of the group to be adopted. We show in the course of the paper that this approach offers numerous advantages, and can lead to both increased reliability and efficiency.
We focus in this paper on providing \emph{computable and efficient} structures for belief aggregation and propagation since good decision making needs to be done at operation time. 
We show in Sec.~\ref{subsec:def} that we rely on \emph{lattices} as formal representation of partial ordering to model ranking of autonomous systems, from more trustworthy to less trustworthy, based on the quality of their attributes. We formally define as well metrics on quality and trustworthiness. 
We further use \emph{Binary Decision Diagrams (BDDs)}~\cite{bdd-seminal} as formally defined structures for belief aggregation and propagation for automated collaborative decision making.
We formulate using the proposed ranking 
an ordering of variables in the BDDs 
and propose reduction rules for efficient beliefs propagation. We later define a metric to quantify good decision making based on reliability, an error is defined as an incorrect belief (e.g., due to a misclassification in the perception pipeline). 
We will see in Sec.~\ref{sec:exp} that increasing reliability with collaboration is not always possible and heavily depends on the quality of the group and applied aggregation rules. 

\paragraph{\textbf{Contribution}}we structure the contribution of this paper in the following manner, 
\begin{itemize}
     \item Considering concept of social epistemology, we define and formalize in Sec.~\ref{subsec:def} the structure of autonomous systems, a notion of expertise and ranking based on quality attributes to define which autonomous system is trustworthy and contributes to belief aggregation and propagation. 
         
    \item we formalize and define a model for belief aggregation and propagation based on binary decision diagrams and well-defined rules in Sec.~\ref{subsec:rules}.  We later present Sec.~\ref{subsec:propag} the notion of BDDs reduction resulting from the resolution of peer disagreement. The notions of collaborative trust and related reliability we use to quantify the results of our approach are  defined in Sec.~\ref{sec:collaborative_reliability}.

    \item Experimental evaluation and validation is presented in Sec.~\ref{sec:exp}. We study the effect of our approach considering different rules and evaluate how reliability increases considering the collaboration. We investigate in particular how the quality of the group influences the overall results, that is the choice of the group of interest which will contribute to good decision making.

   \item we discuss in Sec.~\ref{sec:rw} relevant related work and conclude in Sec.~\ref{sec:concl}  with a discussion about the key outcomes of our proposed approach. 

\end{itemize}

\section{Model of Collaborative Autonomous Systems}
\label{sec:contrib1}
\subsection{Dominance, Trust and Expertise}
\label{subsec:def}
Autonomous systems are different w.r.t. the quality of their attributes, we use this criteria as a mean to define relationships considering the notion of \emph{dominance}. 


\begin{definition}[Autonomous System] 
\label{def:as}
    An autonomous system $s_i = (P_i, \gamma_i)$  is defined by a sphere of  \emph{individual knowledge} $P_i$ and a set of \emph{attributes} $\gamma_i$. 
    The sphere of individual knowledge $P_i \subseteq \mathcal {P}$ can be defined as a the subset\footnote{We make this distinction since an autonomous system can only have a partial knowledge about the environment of operation.}
    $P_i$ of predicates from a larger set $\mathcal {P}$ characterizing an environmental situation. 
     We define the set $\gamma_i$  of attributes as a set of parameters characterizing an autonomous system.  We consider attributes that can be quantified and that can be comparable, so that a ranking based on the quality of attributes can be established. 
    \label{def:autonomous_system}
\end{definition}



\begin{figure}
    \centering
    \includegraphics[scale = 0.65]{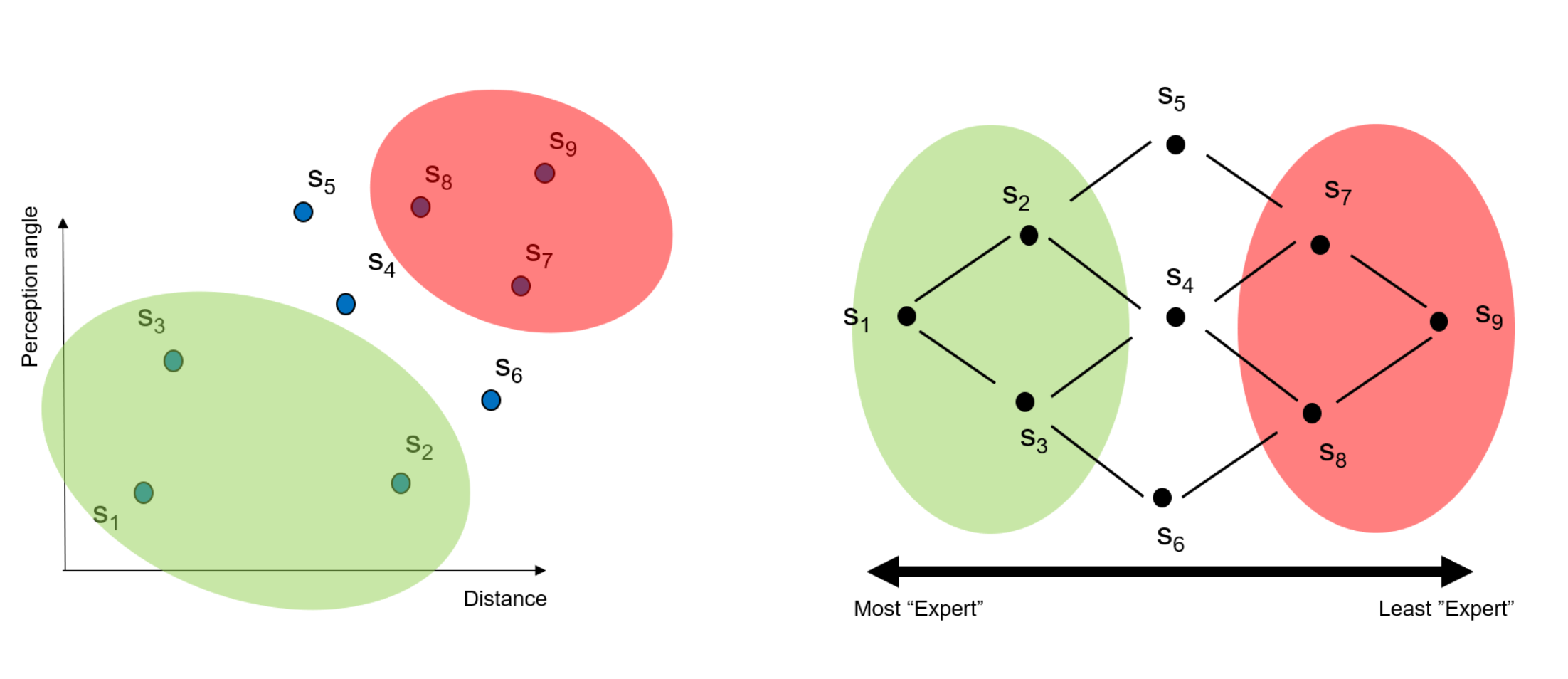}
    \caption{
        Lattice as a formal representation of partial-order relationship between autonomous systems based on their attributes. Green (resp., red) area represent autonomous systems with better (resp., worse) quality of attributes than $s_4$.
    }
    \label{fig:lattice}
\end{figure}

We consider a set $\mathcal{S}$ of autonomous systems that can formulate beliefs on the same set of predicates (i.e., the intersection of their sphere of knowledge is not empty). 
Considering ranking of autonomous systems based on the quality of their attributes, a notion of "expertise" can be defined to qualify autonomous systems of a better quality. For every autonomous system with a given quality of attributes (e.g., $\{s_4\}$ in Fig,~\ref{fig:lattice}), experts are the set of points whose quality dominate its quality for every attribute (e.g., $\{s_1, s_2, s_3\}$). Other autonomous systems whose quality of attributes are dominated by this point can be referred to as less experts. 
We formulate that structure using a \emph{lattice} as a formal representation for partial ordering, with an infimum and a supremum that constitute autonomous systems with worst and best quality of attributes respectively, see Fig~\ref{fig:lattice}.


\begin{definition}[Attributes and Dominance] \label{featuresanddominance}
 Attributes are defined as a d-dimensional vector $(\gamma_1,.., \gamma_d )$
    whose values are ranging over a bounded hypercube $\Gamma$\footnote{Since the quality of attributes can be quantified, we assume a min and max value on the quality of each attribute bounding the set of values a quality of an attribute can have.}. Based on the quality of attributes, we establish an order between autonomous systems $s \in \mathbb{S}$. 
    The set $\mathbb{S}$ is therefore a \emph{lattice} with a partial-ordered set of autonomous systems defined as follows, 
    $$s' \geq s \equiv \forall_i~ \gamma'_i \geq \gamma_i$$
    A strict order relation between autonomous systems can be defined as follows, where the autonomous system $s'$ is better or equal than autonomous system $s$ in \emph{all} attributes and there \emph{exist} an attribute $j$ where $s'$ \emph{dominates} $s$. 
    $$s' > s \equiv  s' \geq s \land \exists_j~ \gamma'_j > \gamma_j$$
    
    Pairs of points such that $s' \not\geq s$ and  $s \not\geq s'$ are \emph{incomparable} (i.e., autonomous systems $s'$ is better than $s$ in at least one attribute and worse in at least another attribute) and are denoted as $s' \parallel s$. 
    
    The meet $s_{inf}$ and join $s_{sup}$ of the lattice $\Gamma$ 
    are defined as follows, and determine respectively the most and least expert in the group of autonomous systems,

    $$s_{inf} = 
    (min\{\gamma_1,\gamma'_1\},...,min\{\gamma_n,\gamma'_n\})$$
    
    $$ s_{sup} = 
    (max\{\gamma_1,\gamma'_1\},...,max\{\gamma_n,\gamma'_n\})$$

\end{definition}

\begin{definition}[Trust and Expertise]
Trustworthiness can be defined as a relationship between two autonomous  $s$ and $s'$ following the dominance relationship, that is $s$ \emph{trusts} $s'$ if $s' \geq s$ and $s'$ is considered as expert by $s$. 

    
\end{definition}





\subsection{Beliefs Aggregation and Peer Disagreement}
\label{subsec:rules}
Every autonomous system can formulate beliefs about predicates in their own sphere of knowledge. In addition to an own belief, every autonomous system receives beliefs from other autonomous systems. 
One of the main issues in the epistemic sense is to deal with \emph{peer disagreement}, the goal of the aggregation function is to reduce errors by adopting beliefs from trustworthy peers. 

\begin{definition}[Belief]
    A belief from an autonomous system $i$ is a (Boolean) statement on a given predicate (i.e., description of a given situation in the environment). 
    The set of beliefs can be defined as a mapping from the set of autonomous systems $\mathcal{S}$ and the set of environment knowledge predicates $\mathcal{P}$ to the set of Booleans, that is $\mathcal{X}: \mathcal{S} \times\mathcal{P} \rightarrow \{True, False\}$. 
    The term $x_i(p)$ defines an 
    assignment from an autonomous system $s_i \in \mathcal{S}$ to a predicate $p \in P_i $ belonging to its sphere of knowledge. 
    \label{def:belief}
\end{definition}

\begin{definition} [Group of Interest]
 A group of interest $G_i$ determines a (sub)group of $k$ peers of autonomous systems that can contribute to forming beliefs for an autonomous system $i$. They can therefore as well contribute to collective decision making for $i$. Every autonomous system $j \in G_i$ belonging to the group of interest has a belief $y_j(p)$ propagated to $i$. Every autonomous system receives beliefs from other autonomous systems belonging to a group of interest. A group of interest can be defined following different rules. In the "Majority" rule,  all the group of autonomous systems contribute to the belief of each member of the group. The group of interest for every autonomous system then becomes the entire set, $ \forall~ s_i, G_i = \mathcal{S}$.  When an expert-based rule is used, the group of interest refers to a (sub)set of experts. In the "Most Expert" rule, only the most expert autonomous system (i.e., the autonomous system with the best quality of attributes) contributes to the belief of each member of the group. The group of interest is then the join of the lattice, $ \forall~~ s_i, G_i = \{s_{sup}\}$. In the "n-expert" rule, for each autonomous system $s_i$, the set of $n$ autonomous system that have better quality of attributes than $s_i$ (i.e., $s_j$  that dominate $s_i$) contribute to its belief, 
 $\forall~~ s_i, G_i = \{s_1, s_2, ..., s_n\} \text{~s.t~~} s_1 > s_i, s_2 > s_i \text{ and } s_n > s_i
$
\end{definition}

\begin{figure}
	\centering
  \includegraphics[scale = 1.6]
{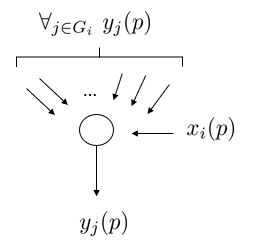}
	\caption{A belief aggregation function considers $k$ input beliefs $y_j(p)$ on a predicate $p$. Belief of a node $x_i(p)$ can change (i.e., $ y_j(p) = \neg x_i(p)$  ) if there is a disagreement with input beliefs.
 }
    
	\label{fig:aggregate_function}
\end{figure}

\begin{definition} [Belief Aggregation]
    An aggregation function $\odot$ is applied by every autonomous system to form an output belief. The function takes as input an own belief $x_i(p)$  and further beliefs from the selected group of interest $G_i$ to compute an output belief $y_i(p)$ that will further be propagated, see Fig.~\ref{fig:aggregate_function}. Considering a group of interest of size $k$, the aggregate function can be defined as a mapping over the set of Booleans $\mathcal{B}$, so that $\odot: \mathcal{B} x \mathcal{B}^k \rightarrow \mathcal{B}$, that is 
    for every belief $x_i(p)$ of $i$ and $k$-tuple of beliefs $y_j(p), j \in [1..k]$ from all autonomous systems in $G_i$, 
    an output belief $y_i(p)$  is computed, 
    
    \begin{equation}
    \label{eq:belief-aggregation}
        y_i(p) =  \odot (x_i(p), y_i^1(p), ..., y_i^k(p))    
    \end{equation}
   
\end{definition}

\begin{definition}[Peer Disagreement]
    Peer disagreement occurs whenever two autonomous systems $(i,j)$ have different beliefs about the same predicate $p$, that is 
$$\exists(i,j) ~~\mbox{s.t.,} ~~x_i(p) \neq y_j(p)$$
\end{definition}
The goal of the aggregation function is to 
reduce peer disagreement, where an autonomous system can change its own belief to adopt the belief of the rest of the group of interest, i.e., for instance the opinion of more experts. By that, we aim at reducing errors (e.g., from one autonomous system perception system) by relying on beliefs from other (trustworthy) members of the group.

Several aggregation rules can be defined, we use in the following a simple aggregation function, 
\begin{equation}
\label{eq:agreement}
 y_i(p) = \odot (x_i(p), ~~\forall j\in G_i~y_i^j (p) )= \left\{
    \begin{array}{ll}
        \neg x_i(p) & \mbox{if } x_i(p) \land \neg y_i^1(p)\land ...\land \neg y_i^k(p) \\
         x_i(p) & \mbox{otherwise.}
    \end{array}
\right.   
\end{equation}

According to this rule, an autonomous system $i$ changes its own belief on a predicate $p$,  $x_i(p)$ when this latter disagrees with all input beliefs $y_j(p)$ belonging to the group of interest for $i$, $G_i$. Since we deal with Boolean variables the output $y_i(p)$ is also Boolean and can be either equal to $x_i(p)$ or $\neg x_i(p)$.

\subsection{Modeling Beliefs Propagation and Reduction}
\label{subsec:propag}
\begin{definition}[Belief Propagation]
Belief propagation is defined recursively by applying the belief aggregation function $\odot$ defined in Eq.~\ref{eq:belief-aggregation}, first to each autonomous systems in the group of interest $s_j \in G_{k}$ to form beliefs $y_i^j(p)$ and then further to autonomous system $s_i$ to form $y_i(p)$ aggregated considering own belief $x_i(p)$. To simplify the notation, we refer in the following to $y_i^j(p)$ directly as $y_j(p)$. Given the n-expert rule, input considered beliefs $y_{i-1}, y_{i-2}, ..., y_{i-k}$ are belonging to the group of interest $G_{i}$   

\begin{equation}
\left\{
    \begin{array}{ll}
     y_j(p) = x_j(p)    \qquad \forall j \in G_{k} 
     \\   
    y_n(p) = \odot (x_n(p), ~ y_{n-1}(p), ~y_{n-2}(p), ..., ~ y_{n-k}(p) )   \qquad \forall n \geq k \\        
    \end{array}
\right.
\end{equation}
\end{definition}

\begin{figure}
	\centering
  \includegraphics[scale = 0.75]
{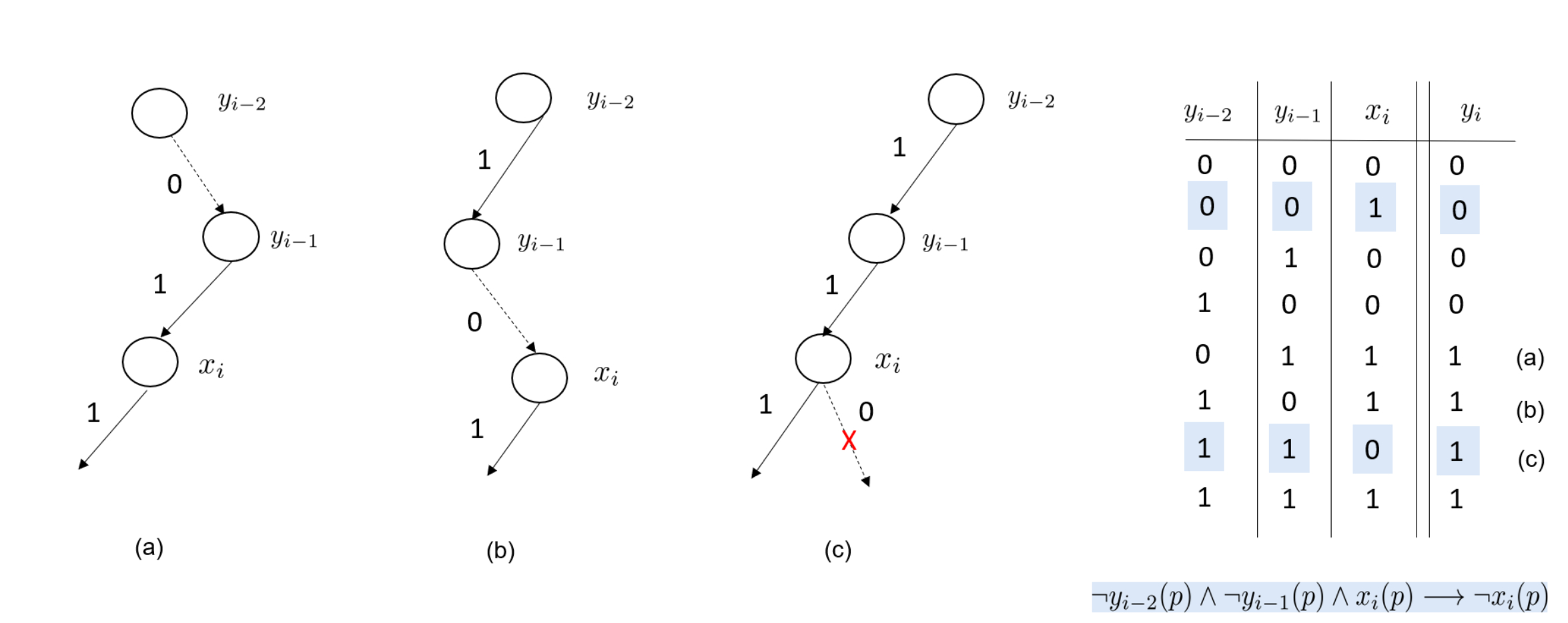}
	\caption{Example of a 2-expert aggregate function for belief aggregation. The function considers for $x_i(p)$ two input beliefs $y_{i-1}$ and $y_{i-2}$ on a predicate $p$ to compute $x_i(p)$.  Belief of a node $x_i(p)$ is changed if there is a disagreement with all input beliefs $y_j(p)$ .}
	\label{fig:belief_transformation}
\end{figure}

The effect of belief aggregation is depicted in Fig.~\ref{fig:belief_transformation}, considering a 2-expert rule where $s_{i-2}$ and $s_{i-1}$ dominate $s_i$ (i.e., having better quality of attributes), and contribute in case of peer disagreement, using the aggregation function, to a change of belief of $s_i$. The aggregation function 
considering the 2-expert rule is boolean 
and is defined by the truth table depicted in Fig.~\ref{fig:belief_transformation}, considering all possible combinations of beliefs aggregation. 
According to the defined 2-expert rule function, a change of belief of $s_i$(i.e., $y_i(p) =  \neg x_i(p)$) occurs when it contradicts the belief of \emph{both} experts (i.e.,  $\neg y_{i-2}(p) \land \neg y_{i-1}(p) \land  x_i(p) \rightarrow \neg x_i(p) $) 
, otherwise the belief remains unchanged (i.e., $y_i(p) = x_i(p)$). 
Note that, the order of propagation  $y_{i-2}(p)$,  $y_{i-1}(p)$ and $x_{i}(p)$ of some combinations (i.e., (a), (b) and (c)) is depicted considering paths of propagation. 
In the following, we use Binary Decision Diagrams (BDDs) as generalized structured representation of boolean aggregation and propagation functions with all possible paths. We will later apply ordering and reduction rules for beliefs propagation that lead to efficient performance when beliefs propagate to the rest of the group.  


\subsubsection{Beliefs Propagation as Ordered Binary Decision Diagrams (OBDDs)}


Binary decision diagrams represent boolean functions as rooted acyclic graphs, see Fig.~\ref{fig:bdd_reduction}-(a).  Every non-terminal vertex is labeled by a variable for an autonomous system $i$ and represents its belief $x_i(p)$. Every vertex has then two arcs directed towards children nodes, the first arc (shown as a dashedline) corresponds to $x_i(p) =0$
, and the second arc corresponds to the case where $x_i(p) =1$. 
Each terminal vertex is labeled 0 or 1 and represent the belief of the last autonomous system in the group. 

Variables in the BDD can be represented in any order. We use the lattice as a reference model of autonomous systems based on their quality of attributes to define an Ordered Binary Decision Diagram (OBDDs). 
The adopted order of variables in the OBDDs structure follows therefore a total order from the lattice. In this case, beliefs propagation in the BDDs starts from the most expert autonomous system to the least expert one. The aggregation function is used by every node in the BDD to decide whether the autonomous system changes its own belief to adopt the belief of the most expert one(s). The aggregated belief is further propagated and contributes later to beliefs aggregation. 
Therefore for every vertex $y_{i-1} (p)$ and a non-terminal child $y_{i} (p)$, their respective variables must be ordered, that is $y_{i-1} (p) < y_{i} (p)$. This order follows the order depicted in the lattice, that is 
$$s_{i-1} < s_i \implies y_{i-1} (p) < y_{i} (p)$$

\subsubsection{Belief Reduction and Convergence of Beliefs}
\begin{figure}
	\centering
  \includegraphics[scale = 0.5]
{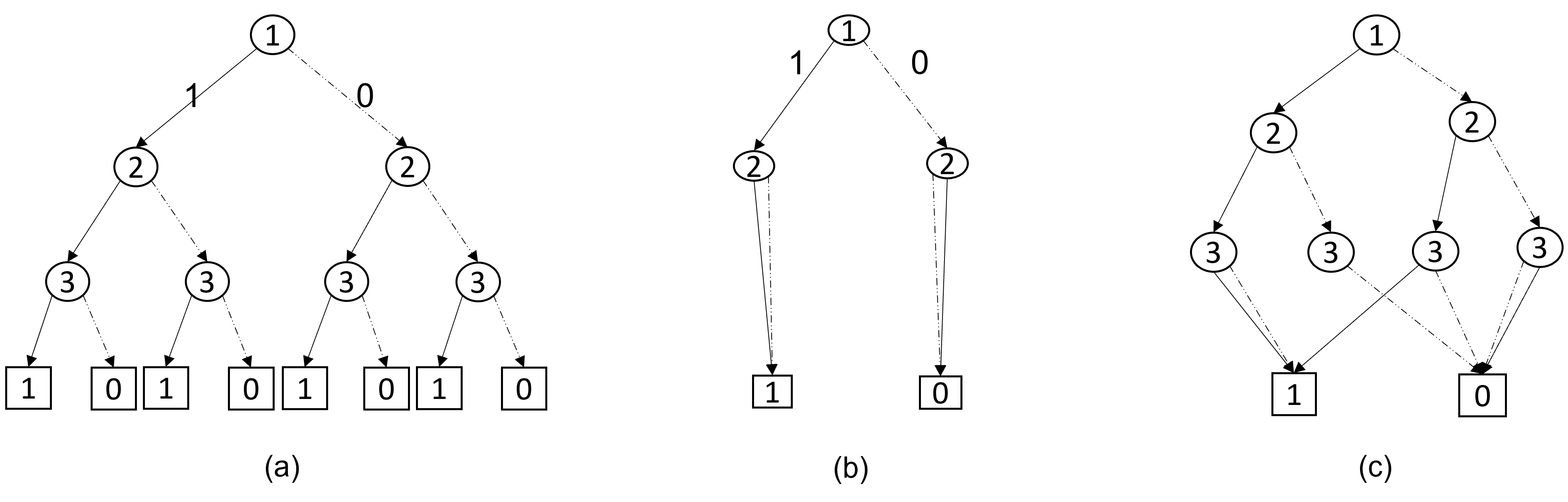}

\caption{(a): non reduced OBDDs with three autonomous systems $s_1 > s_2 > s_3$ 
with order from the most expert to the least expert one, (b): reduced BDD after aggregation and propagation and their beliefs $y_i(p)$ considering the most expert rule where 
all remaining autonomous systems 
adopt the belief of the most expert one, (c) reduced BDD after aggregation and propagation considering the 2-experts rule where paths are reduced by adopting the belief of the two most expert ones when these latter have the same belief.}


	\label{fig:bdd_reduction}
\end{figure}

Belief aggregation and propagation lead to a reduction of the BDD as depicted in Fig.~\ref{fig:bdd_reduction}-(b) and -(c), where less experts autonomous systems adopt the belief of the most expert one (s). We define the following transformation rules over the defined BDDs, allowing reduction and efficient beliefs propagation.  
\begin{enumerate}
    \item Remove duplicate terminals: as in~\cite{bdd-seminal}, eliminate all but one terminal vertex with a given label and redirect all arcs into the eliminated vertices to the remaining one. 
    Following our proposed model terminal vertices correspond to the belief of the last autonomous system(s). 

    \item For each path in the graph, the aggregation function $\odot$ is applied at every node $x_i(p)$ considering as input beliefs as assignment along the path until node $i$ to form the belief  $y_i(p)$.  At every node, $x_i(p)$ is then replaced by $y_i(p)$ after belief aggregation. 


    \item the aggregated belief $y_i(p)$ is further propagated and contributes to beliefs aggregations of other autonomous systems along the rest of the paths.    
\end{enumerate}

Aggregation and propagation rules lead to reduction of the binary diagram where \emph{less experts adopt the belief of most expert one(s)}, see Fig.~\ref{fig:bdd_reduction}-(b),-(c). According to the aggregation rule defined in Eq~\ref{eq:agreement}. If we have a sequence of $k$ similar beliefs, that is $[\bar{n} -k,\bar{n}]$ so that $y_{\bar{n}-k} = ... = y_{\bar{n}}$, then all beliefs $n \geq \bar{n}$ will converge. Convergence here means also that peer disagreement  will disappear after the propagation, that is 

\begin{equation}
\label{eq:convergence}
  \forall n \geq \bar{n},~~ y_{n}  = y_{\bar{n}} = ... =  y_{\bar{n}-k}  
\end{equation}

We can prove that by induction as follows, 
if we assume that \ref{eq:convergence} is true at step $n$, this means that $y_{n-1}(p)= y_{n-2}(p)=... = y_{n-k}(p)$ and as a consequence $y_{n}(p) = y_{n-1}(p)= y_{n-2}(p)=... = y_{n-k}(p)$.

Now let us prove \ref{eq:convergence} for step $n+1$, that is we need to prove that  $y_{n+1}(p) = y_{n}(p) = y_{n-1}(p)= y_{n-2}(p)=... = y_{n-k}(p)$.  According to the rule in Eq.~\ref{eq:agreement}, $ y_{n+1}(p) = x_{n+1}(p) \neq y_{n-1}(p), ...$ when there exist one $i,j$ so that $ y_{n+1}^{i} \neq  y_{n+1}^{j} $, in this case, according to the rule $i$ does not change its own belief and $ x_i(j) = y_{n+1}^{j}$  or $x_i(j) = y_{n+1}^{j}$. Now we know from the previous iteration that beliefs from $y_{\bar{n}-k} = ... = y_{\bar{n}}$ agree (this is our assumption) and we know as well that $y_{n+1}^{i}$ also agrees with the previous $k$ based on rule \ref{eq:agreement} (either belief was the same as the previous k or just changed). In this case, we know that all beliefs $y_{\bar{n}-k} = ... = y_{\bar{n}}, y_{\bar{n}+1 }$ agree which will lead to a convergence of beliefs.

Note that, the convergence depends on the applied aggregation rule and allows to determine the efficiency of belief aggregation and propagation. 
\section{Good Decision Making and Collaborative Reliability}
\label{sec:collaborative_reliability}


\begin{definition}[Correctness of a Belief and Errors]
A belief 
on a given predicate $p$ is correct if it matches the ground truth. We assume so far for the considered model that the ground truth $T$ is known. 
Since we rely on boolean algebra, an \emph{error} $\uptheta_{i,p} $ of a belief 
from an autonomous system $i$ on a predicate $p$ can be defined as a 
boolean function equal to $1$ (i.e., true) when predicates $x_i (p)$ and $T(p)$ have different (true or false) values, otherwise it is equal to $0$ (i.e., false). The boolean function $\uptheta_{i,p} $ can then simply be viewed as a XOR function. 
The boolean function defining \emph{Correctness} $\upzeta_{i,p}$ is the inverse function defining errors and can be defined as well as a boolean function equal to $1$ (i.e., true) when beliefs $x_i (p)$ and $T(p)$ are similar equal both to true or false, the function is equal to $0$ otherwise when $x_i (p)$ is different from the ground truth  $T(p)$.

\begin{equation}
\label{eq:error-correct}
 \left\{
    \begin{array}{ll}
       \upzeta_{i,p} = x_i (p) \oplus T(p)\\ 
      ~~\\
      \uptheta_{i,p} = \neg \upzeta_{i,p} =  \overline{x_i (p) \oplus T(p)}
    \end{array}
\right.
\end{equation}

\end{definition}

%
%


\begin{definition}[Reliability]
Reliability $R_i$ of an autonomous system $i$ before the collaboration is defined solely based on $x_i (p)$ modeling its sphere of knowledge and (individual) errors (resp., correctness). Reliability of an autonomous system can then be defined as a ratio over all predicates 
$p\in P$, of the sum of correct beliefs over the total sum of all (correct and erroneous) beliefs. 
It can be defined as follows, 
\begin{equation}
\label{eq:individual_reliab}
    R_i = \sum_{p\in P} \frac{ \upzeta_{i,p} }{ \upzeta_{i,p} +\uptheta_{i,p}  } 
\end{equation}
\end{definition}

\begin{definition}[Collaborative Reliability]
\label{sec:col_rel}
When using the collaboration, a change of belief may occur, that is $y_i(p) \neq x_i(p)$.
Therefore, both correctness and error as defined by Eq.~\ref{eq:error-correct} become defined 
based on the output belief $y_i (p)$ after belief aggregation and propagation. 
 we refer to this correctness (resp., error) considering a possible change of belief after the collaboration as $\upzeta_{i,p}^c$ (resp., $\uptheta_{i,p}^c$ ) and is then defined as follows, 

%
\begin{equation}
 \left\{
    \begin{array}{ll}
      \upzeta_{i,p}^c = y_i (p) \oplus T(p) \\ 
      ~~\\
      \uptheta_{i,p}^c = \neg \upzeta_{i,p}^c =  \overline{y_i (p) \oplus T(p)}
    \end{array}
\right.
\end{equation}
\end{definition}
%
%
%
%
Due to the collaboration and the aggregation function, some autonomous systems adopt the beliefs of others autonomous systems which are more experts (when the n-expert rule is used), thereby discarding their own belief. 
 If the change of belief leads to $y_i(p) = T(p)$, the belief is corrected. In the opposite case, $y_i(p) \neq T(p)$ the belief is not corrected, but rather an error from most experts has been propagated using the belief propagation and aggregation function. In the case where $y_i(p) = x_i(p)$ the belief has not been modified by belief propagation and aggregation, thereby resulting in similar number of errors. In order to distinguish between the aforementioned cases, we distinguish after the collaboration between the following
\begin{itemize}
    \item Individual errors: errors that have occurred before the collaboration and have been maintained after the collaboration (i.e., have not been corrected after applying the rule), they are defined when $y_i(p) = x_i(p)$ and $\uptheta_{i,p} = 1$. 

    \item Introduced Errors: errors that have been introduced by the collaboration after applying the rule and where errors of most experts autonomous systems have propagated and led to a change of a correct belief, they are defined when $y_i(p) \neq x_i(p)$  and $\uptheta_{i,p}^c = 1$.

    \item Corrected Errors: errors that have been corrected by the collaboration after applying the rule and where correct beliefs of most experts autonomous systems have propagated and led to a change an erroneous belief, they are defined when $y_i(p) \neq x_i(p)$  and $\upzeta_{i,p}^c = 1$.
\end{itemize}
We therefore define collaborative reliability $R_i^c$ as a metric over all predicates of the ratio between unchanged individual errors ($\sum_{\forall_{i,p}}^{x_i(p) = y_i(p)}~ \upzeta_{i,p}$) and additional introduced errors after the collaboration ($\sum_{\forall_{i,p}}^{x_i(p) \neq y_i(p)} ~ \uptheta_{i,p}^c$) over the number of unchanged individual errors and number of corrected errors after collaboration ($\sum_{\forall_{i,p}}^{x_i(p) \neq y_i(p)} ~\upzeta_{i,p}^c$). 
When the number of introduced errors and corrected errors is the same then the collaborative reliability ratio is equal to $R_i^c = 1$. When all errors are corrected and no error is introduced, then the collaborative reliability ratio is optimally equal to $R_i^c = 0.5$. Inversely, when the number of introduced errors is higher than the number of corrected errors then $R_i^c \geq 1$.




\begin{equation}
\label{eq:collab-reliab}    
R_i^c = \sum_{p\in P}  
\frac{\sum_{\forall_{i,p}}^{x_i(p) = y_i(p)}~ \upzeta_{i,p} + \sum_{\forall_{i,p}}^{x_i(p) \neq y_i(p)} ~ \uptheta_{i,p}^c }{\sum_{\forall_{i,p}}^{x_i(p) = y_i(p)} ~ \upzeta_{i,p} + \sum_{\forall_{i,p}}^{x_i(p) \neq y_i(p)} ~\upzeta_{i,p}^c}
\end{equation}



\begin{table}
\centering
\begin{tabular}{ |c|c|c|c|c|c|c|c|c| } 
\hline
Config. \# & 
$x_i (p)$ & 
$T(p)$ & 
$\upzeta_{i,p}$ & 
$\uptheta_{i,p}$ &$\upzeta_{i,p}^c$ & $\uptheta_{i,p}^c$ & Error Status\\
\hline
1 & 1 & 1 & 1 & 0 & \textbf{0} & \textbf{1} &  introduced\\ 
2 & 1 & 0 & 0 & 1 & \textbf{1} & \textbf{0} &  corrected\\ 
3 & 0 & 1 & 0 & 1 & 0 & 1 &  unchanged\\ 
4 & 0 & 0 & 1 & 0 & 1 & 0 &  unchanged \\
\hline
\end{tabular}

\caption{Example of different configurations, each configuration constitute a possible assignment of beliefs $x_i(p)$ and their correctness or not w.r.t. the ground truth $T(p)$. Introduced and corrected errors after the collaboration are shown.}
\end{table}

\begin{table}[H]


\begin{adjustbox}{width=1\textwidth}
\begin{tabular}{|c |c|}
\hline

\textbf{Parameter}      &      \textbf{Description} 
\\ \hline

$s_i$  & autonomous system $i$
\\ \hline

$s_i \geq s_j $  & quality of attributes of autonomous system $i$ is better or equal than the quality of attributes of autonomous system $j$
\\ \hline

$x_i(p)$  & Belief of autonomous system $i$ regarding a proposition $p$ \emph{before} belief propagation
\\ \hline

$y_i(p)$  & Belief of autonomous system $i$ regarding a proposition $p$ 
 \emph{after} belief propagation
\\ \hline

$G_i$  & The set of autonomous system whose belief will propagate to $s_i$ to contribute to forming the belief $y_i$  
\\ \hline

$\odot$  & Beliefs aggregation function,  $y_i(p) = \odot (x_i(p), ~~\forall j \in G_i ~y_j(p))$
\\ \hline

$\upzeta_{i,p}$ & Correctness of a belief $x_i(p)$ from an autonomous system $i$ considering a proposition $p$
\\ \hline

$\uptheta_{i,p} $ & error of a belief $x_i(p)$ from an autonomous system $i$ considering a proposition $p$
\\ \hline

$\upzeta_{i,p}^c$ & Correctness of a belief $y_i(p)$ from an autonomous system $i$ considering a proposition $p$
\\ \hline

$\uptheta_{i,p}^c $ & error of a belief $y_i(p)$ from an autonomous system $i$ considering a proposition $p$
\\ \hline

$R_i $ & Reliability of an autonomous system $i$ based on errors performed individually before the collaboration
\\ \hline

$R_i^c $ & Collaborative Reliability of an autonomous system $i$ based on errors performed individually, errors corrected and errors introduced after the collaboration. 
\\ \hline

\end{tabular}
\end{adjustbox}
\caption{Overview of all notations and respective description used in Sec.~\ref{sec:contrib1} and Sec.\ref{sec:collaborative_reliability}.}

\end{table}

\section{Experiments and Validation}
\label{sec:exp}

\begin{figure}
    \centering
    \begin{subfigure}[b]{0.39\textwidth}
        \centering
        \includegraphics[width=\textwidth]{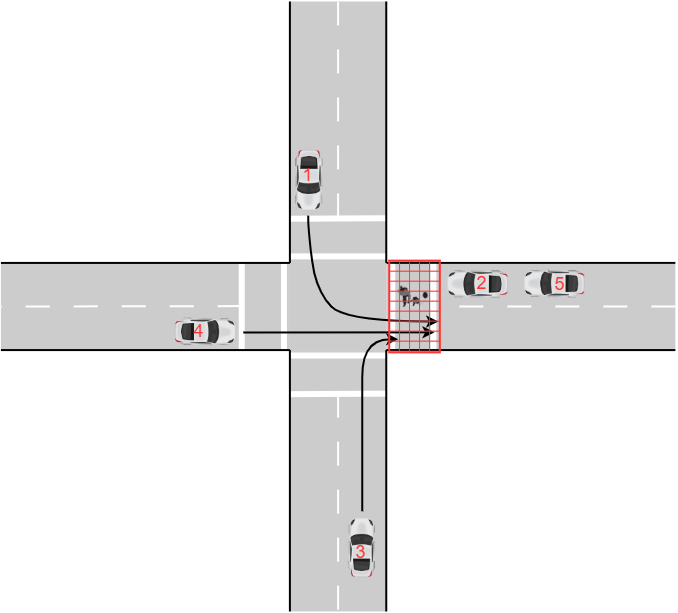}
        \caption{}
    \label{fig:intersection_without_perception}
    \end{subfigure}
    \hspace{1cm}
    \begin{subfigure}[b]{0.39\textwidth}
        \centering
        \includegraphics[width=\textwidth]{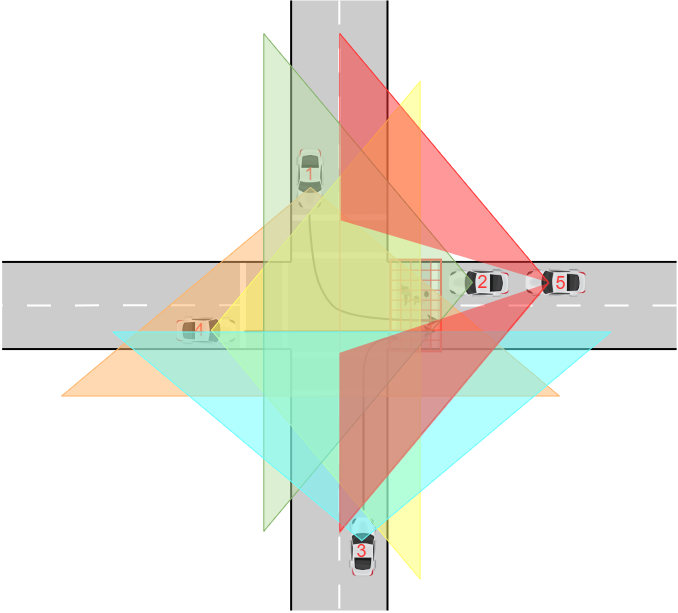}
        \caption{}
        \label{fig:intersection_setup}
    \end{subfigure}
%
        
    \caption{ A four-way intersection with five vehicles with individual FOV. The FOVs indicate object detection capabilities within the ROI. The grid overlay quantifies perception coverage of ROI—the pedestrian crosswalk, while the target object size of pedestrians indicates perception accuracy. 
    }
    \label{fig:intersections}
\end{figure}

In order to illustrate the concepts of this paper, we first consider for evaluation a similar example as in Fig.~\ref{fig:example} as a representative practical use case of a four-way intersection. Fig.~\ref{fig:intersections}-(\subref{fig:intersection_without_perception}) 
presents 
five autonomous vehicles positioned at different locations, each with a distinct color-coded 100-degree field of view (FOV) extending outward from their front. The predicate incorporated within the region of interest (ROI) is: "The pedestrian crosswalk is free", 
a vehicle will either perceive the crosswalk as free (logic 0) or detect a pedestrian (logic 1). 
The FOV of each vehicle determines their ability to detect objects, particularly within the ROI, which is the pedestrian crosswalk. A grid overlay is present in the intersection, serving as a means to quantify perception coverage of the ROI by counting the number of grid squares each vehicle’s FOV covers. This quality attribute determines how well a vehicle perceives the ROI, i.e., the more squares covered, the better the autonomous vehicle’s attribute is. The second quality attribute being measured is the target object size in the frame of the vehicle’s camera sensor, where the target is the pedestrian in our case. The target object size is the number of pixels an object occupies in the image captured by the camera sensor and this is determined by the distance between the vehicle and the pedestrian~\cite{hecht2023optik}, as closer proximity results in a larger object size representation in the captured image. For this experiment, we assume that all vehicles are equipped with identical cameras, ensuring that the camera parameters are the same across all vehicles. A higher target object size generally enhances detection accuracy, making this attribute essential for evaluating how well each autonomous system can visually perceive and interpret the pedestrian's presence within the intersection. 
Table.~\ref{fig:combined_table_lattice}-(\subref{tab:vehicle-attributes}) 
summarizes each vehicle individual belief together with its quality attributes.
Ranking of vehicles based on the defined quality attributes is illustrated in Fig.~\ref{fig:combined_table_lattice}-(\subref{fig:lattice_practical_example}). Vehicles 4 and 1 are incomparable considering both quality attributes, since vehicle 4 has better coverage of the ROI, while vehicle 1 has a better target object size.

\begin{figure}
    \centering
    \begin{subfigure}[b]{0.3\textwidth} 
        \centering
        \includegraphics[height=5cm]{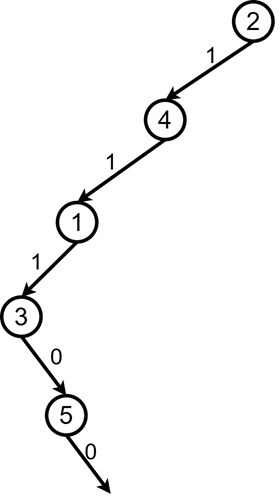}
        \caption{}
        \label{fig:path1}
    \end{subfigure}%
    \hfill 
    \begin{subfigure}[b]{0.3\textwidth} 
        \centering
        \includegraphics[height=5cm]{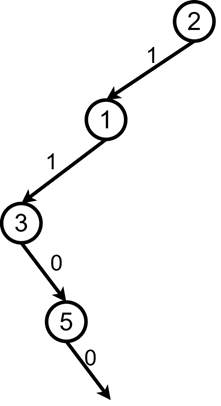}
       \caption{}
        \label{fig:path2}
    \end{subfigure}%
    \hfill 
    \begin{subfigure}[b]{0.3\textwidth} 
        \centering
        \includegraphics[height=5cm]{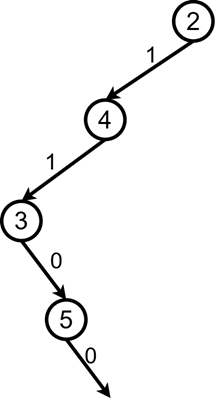}
       \caption{}
        \label{fig:path3}
    \end{subfigure}

    \caption{
    Corresponding paths extracted from the OBDD for the autonomous vehicles, determined by the beliefs of each vehicle and ranking of quality attributes.} 
    \label{fig:combined_paths}
\end{figure}

For different paths in the lattice, i.e., a total order of autonomous vehicles, a corresponding ordered BDD can be constructed following the order of expertise. 
Fig.~\ref{fig:combined_paths} represents different paths in the obtained OBDD considering beliefs and experts in the present described environmental situation. 
Applying the two-expert rule across all paths corrects the beliefs of vehicles 3 and 5 (from logic 0 to logic 1). However, when using the three-expert rule, only path 1 in Fig.~\ref{fig:combined_paths}-(\subref{fig:path1}) corrects the beliefs of vehicles 3 and 5; this is because the three-expert rule requires three experts to share the same belief for it to be propagated. In paths 2 and 3, this condition is not met. 
In the following, we generalize this example. 

\newcolumntype{C}[1]{>{\centering\arraybackslash}p{#1}}
\begin{figure}
    \centering
    \begin{subfigure}[b]{0.55\textwidth} 
        \centering
        \renewcommand{\arraystretch}{1.3}
         \scriptsize
        \begin{tabular}{|C{0.8cm}|C{0.7cm}|C{1.4cm}|C{2cm}|C{0.9cm}|}
            \hline
            \textbf{Vehicle} & \textbf{Belief} & \textbf{Attribute 1 (target object 
            size)} & \textbf{Attribute 2 (\% of ROI covered by FOV)} & \textbf{Ranking of expertise in the lattice} \\
            \hline
            1 & 1 & 1271 & 53.3 & 2 \\
            2 & 1 & 3766 & 66.7 & 1 \\
            3 & 0 & 748  & 22.2 & 3 \\
            4 & 1 & 915  & 60   & 2 \\
            5 & 0 & 0    & 26.7 & 4 \\
            \hline
        \end{tabular}
        \caption{}
        \label{tab:vehicle-attributes}
    \end{subfigure}%
    \hfill 
    \begin{subfigure}[b]{0.35\textwidth} 
        \centering
        \includegraphics[width=\textwidth]{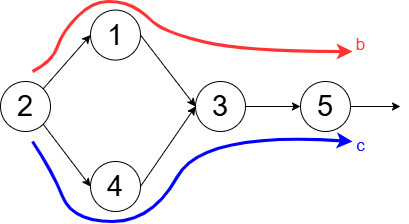}
     \caption{}
        \label{fig:lattice_practical_example}
    \end{subfigure}
    
    \caption{Overview of vehicle attributes and lattice representation. (a) Table displaying the attributes and beliefs of the autonomous vehicles, including target object size and ROI coverage. (b) Lattice illustrating ranking of vehicles based on their quality attributes and multiple paths represents a total order in the lattice. } 
    \label{fig:combined_table_lattice}
\end{figure}

\subsection{Experimental Setup}

We perform further experiments to evaluate enhancing the reliability of autonomous systems through collaboration. 
We generate synthetically a group of autonomous systems, each characterized by different quality attributes. 
For every autonomous system, we create randomly generated individual beliefs, 
which are subsequently aggregated and propagated using various rules to collaboratively correct beliefs. 
For that, we construct ordered binary decision trees for belief aggregation and propagation, with the order determined by the quality of attributes. Using our defined metric of collaborative reliability, 
we assess the effects of collaboration by varying aggregation rules and the quality of the group. 

\subsubsection{Generation of Quality Attributes}
We consider multiple quality attributes for each autonomous system (e.g., distance and perception). Assignment of the quality of attributes to each autonomous system is done randomly in the form of a percentage selected from a uniform distribution between [0..100], where 100\% is the best possible quality for a given attribute. 
 After generating a set of autonomous systems with different quality attributes, we construct 
 a lattice as a partial-order ranking structure. A total order used to obtain an OBDD is based on a  single path of the lattice.


\subsubsection{Generation of Beliefs} We define multiple different predicates $p_1, p_2, ..., \mbox{etc.}$ (e.g., to model statements such as "an object has been detected") and establish the ground truth as the correct assignment of all considered predicates. We generate randomly for every predicate and every autonomous system beliefs in the form of $0$ (i.e., False) or $1$ (i.e, True). For our synthetic benchmarks and when generating beliefs, we assume that the error rate (i.e, the number of times an autonomous system emits an erroneous belief that does not match the ground truth) of autonomous systems with better quality attributes is lower than the error rate of autonomous systems with worse quality attributes. 
Therefore, the error rate for every autonomous systems belief is variable and we assume that it can be calculated using a linear factor based on the quality of the attributes. 
Table ~\ref{tab:benchmark-example-table} provides an example of random assignments of quality of attributes and beliefs following the example in Fig,~\ref{fig:lattice}. 



\subsubsection{Considered Aggregation Rules}
We evaluate our setup considering multiple aggregation rules, namely, the "n-experts" rule with the following variants: the "Most Expert", "Two Experts", "Three Experts", "Four Experts", in addition to the  "Majority" rule, and the "Gravity Point" rule defined below. 
The "Gravity Point" rule considers, similarly to the majority rule, beliefs from both autonomous systems with better and worse quality of attributes. However, it considers, unlike the majority rule, only a subgroup as the group of interest and not all the set of autonomous systems. By considering quality of attributes and their ranking, the considered autonomous system constitutes the gravity point of the group of interest. 

\begin{table}
\centering
\begin{adjustbox}{width=1\textwidth}
\begin{tabular}{ |c|c|c|c|c|c| } 
\hline
Autonomous System $s_i$& 
Quality Attr. $\gamma_1$ (\%)& 
Quality Attr. $\gamma_2$ (\%)&
Config.~1: Belief $x_i (p)$  &
Config.~2: Belief $x_i (p)$ & $T (p)$ \\
\hline
$s_1$  & 95\% & 96\%& 1 & 1& 1\\ 
$s_2$  & 75\%& 95\%& 1 & 1 &  1\\ 
$s_3$  & 93\%& 80\% & 0 & 1 &  1\\ 
$s_4$  & 76\%& 78\% & 1 &1  &  1\\
$s_5$  & 77\%& 69\% & 1 &0  &  1\\
$s_6$  & 70\%& 80\% & 1 &1  &  1\\
$s_7$  & 68\%& 70\% & 0 &1  &  1\\
$s_8$  & 72\%& 69\% & 0 &0  &  1\\
$s_9$  & 50\%& 50\% & 0 &0  &  1\\
\hline
\end{tabular}
\end{adjustbox}
\captionof{table}{Example of a synthetic benchmark where quality of attributes and beliefs for a predicate $p$ are generated randomly for two different configurations. The resulting ranking of autonomous systems is modeled after the lattice in Fig.~\ref{fig:lattice}}.
\label{tab:benchmark-example-table}
\end{table}

\subsection{Effect of the Rules on Collaborative Reliability}
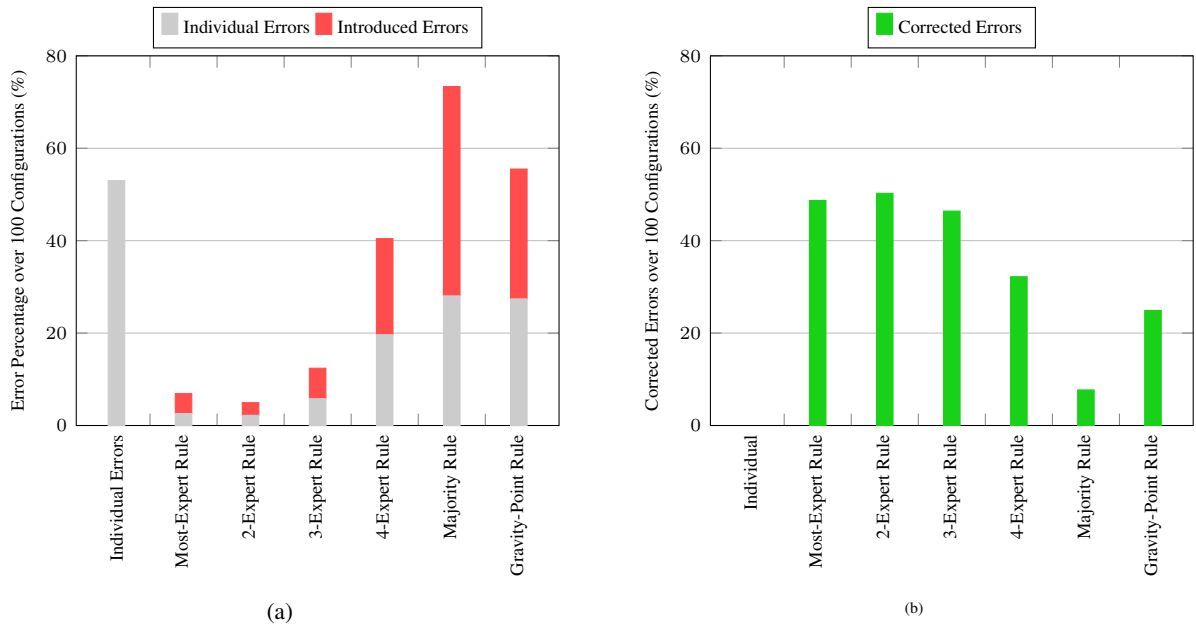
\begin{figure}[H]
\centering
\begin{multicols}{2}
\begin{subfigure}{0.37\textwidth}
\hspace*{-20pt}
\begin{tikzpicture}
\begin{axis}[
    ybar stacked,
    bar width=0.22cm,
    ymin=0,
    ymax=80,
    width=8cm,
    height=6.5cm,
    yticklabel style={font=\scriptsize},
    symbolic x coords={Individual Errors, Most-Expert Rule, 2-Expert Rule, 3-Expert Rule, 4-Expert Rule, Majority Rule, Gravity-Point Rule},
    xtick={Individual Errors, Most-Expert Rule, 2-Expert Rule, 3-Expert Rule, 4-Expert Rule, Majority Rule, Gravity-Point Rule},
    xticklabel style={yshift=0ex, rotate=90, font=\scriptsize},
    x label style={at={(axis description cs:0.5,-0.1)},anchor=north},
    ylabel={Error Percentage over 100 Configurations (\%)},
    label style={font=\scriptsize},
    major x tick style={opacity=0},
    minor x tick num=1,
    minor tick length=1 ex,
    ymajorgrids= true,
    legend style={
    at={(0.50,1.02)},
    anchor=south,
    column sep=0ex,
    legend columns=-1,
    }
]
\addplot [fill=lightgray!80,draw=lightgray!80] plot coordinates {
    (Individual Errors,53) 
    (Most-Expert Rule,2.6) 
    (2-Expert Rule,2.2) 
    (3-Expert Rule,5.8) 
    (4-Expert Rule,19.65)
    (Majority Rule,28.05)
    (Gravity-Point Rule,27.4)
};
\addlegendentry{\scriptsize{Individual Errors}};
\addplot[fill=white!30!red!100,draw=white!30!red!100] plot coordinates {
    (Individual Errors,0)
    (Most-Expert Rule,4.3) 
    (2-Expert Rule,2.75) 
    (3-Expert Rule,6.6) 
    (4-Expert Rule,20.8)
    (Majority Rule,45.3)
    (Gravity-Point Rule,28.1)
};
\addlegendentry{\scriptsize{Introduced Errors}};
\end{axis}
\end{tikzpicture}
    \caption{
    }
    \label{fig:errors1_aft_coll_high_quality}
    \end{subfigure}%
    
    \columnbreak

\begin{subfigure}{0.37\textwidth}
\hspace*{-20pt}
\begin{tikzpicture}
\begin{axis}[
    ybar stacked,
    bar width=0.22cm,
    width=8cm,
    height=6.5cm,
    ymin=0,
    ymax=80,
    yticklabel style={font=\scriptsize},
    symbolic x coords={Individual, Most-Expert Rule, 2-Expert Rule, 3-Expert Rule, 4-Expert Rule, Majority Rule, Gravity-Point Rule},
    xtick={Individual, Most-Expert Rule, 2-Expert Rule, 3-Expert Rule, 4-Expert Rule, Majority Rule, Gravity-Point Rule},
    xticklabel style={yshift=0ex, rotate=90, font=\scriptsize},
    x label style={at={(axis description cs:0.5,-0.1)},anchor=north},
    ylabel={Corrected Errors over 100 Configurations (\%)},
    label style={font=\scriptsize},
    major x tick style={opacity=0},
    minor x tick num=1,
    minor tick length=1 ex,
    ymajorgrids= true,
    legend style={
    at={(0.50,1.02)},
    anchor=south,
    column sep=0ex,
    legend columns=-1,
    }
]
\addplot [fill=black!20!green!90,draw=black!20!green!90] plot coordinates {
    (Individual,0)
    (Most-Expert Rule,48.7) 
    (2-Expert Rule,50.25) 
    (3-Expert Rule,46.4) 
    (4-Expert Rule,32.2)
    (Majority Rule,7.7)
    (Gravity-Point Rule,24.9)
};
\legend{\scriptsize{Corrected Errors}}
\end{axis}
\end{tikzpicture}
    \captionsetup{font=tiny}
    \caption{
    }
    \label{fig:errors_aft_coll_high_quality}
    \end{subfigure}%
    \end{multicols}
    \caption{The effect of rules on the number of introduced errors and  corrected errors, (a) Individual errors before belief aggregation and propagation, and introduced errors caused by each rule, (b) Corrected errors after beliefs aggregation and propagation for each rule.  
    }
    \label{fig:results}
\end{figure}

In the first series of experiments, we use synthetic benchmarks generated considering 20 autonomous systems and two attributes and their qualities.  We generate for every autonomous system, as explained in the previous section, beliefs based on quality of attributes. 
For the same example with a given number of autonomous systems and same quality of attributes, we generate multiple \emph{configurations}, each configuration constitute an assignment of beliefs. 
We apply for every configuration aggregation rules and propagation. In order to obtain sound comparisons between the applied rules, the same generated example is used for each rule to observe the effect of rules when using the collaboration on both errors correction (i.e., number of corrected errors) and error creation (i.e., number of introduced errors). We generate up to 100 different configurations for the synthetic benchmark with 20 autonomous systems, results on the percentage of errors are summarized in Fig.~\ref{fig:results}. 

Individual errors are errors resulting from the quality of attributes of each autonomous determining a given error rate. Individual errors characterize the level of reliability (before the collaboration) that we aim to improve with the collaboration and is used  as a baseline for comparison when applying beliefs aggregation and propagation rules. 
As observed in Fig.~\ref{fig:results}-(\subref{fig:errors1_aft_coll_high_quality}), 
for the considered examples, the percentage of the sum of all individual errors over 100 configurations before collaboration is depicted (52.7\%). This percentage shows all errors from autonomous systems when assigning beliefs (without collaboration) randomly following the error rates over the 100 configurations.  Applying each rule leads to number of introduced errors and a number corrected ones. Percentages of the sum of introduced and corrected errors over all 100 configurations for each rule are depicted in Fig.~\ref{fig:results}-(\subref{fig:errors1_aft_coll_high_quality}). 
We overall observe that by applying the rules, the number of corrected errors is increased by using the collaboration to reach the best values with the most expert and the 2-experts rule as observed in Fig. ~\ref{fig:results}-(\subref{fig:errors_aft_coll_high_quality}). Both as well lead to a small number of introduced errors. 


We observe that the most-expert rule allows to introduce a relatively small percentage (3\%)  of errors compared to the majority rule (50\%). The most-expert rule allows as well to introduce a large percentage of corrected errors (50.4\%) compared to the majority rule (8\%). We also observe that overall the two-experts rule outperforms all other rules resulting in the minimum percentage of introduced errors (2\%) and maximum percentage of corrected errors after beliefs propagation (50.8\%). The 2-experts rule has nearly similar performance compared to the most-expert rule when it comes to corrected percentage of errors. However, the percentage of introduced errors is higher in the case of the most-expert rule, having a (1\%) difference. Indeed in the n-expert rule, errors are introduced if all experts make simultaneously an error. Therefore the 2-expert rule is more robust compared to the Most-expert rule as both experts have to make errors simultaneously which is less likely to happen compared to one expert. The "3-expert" and the "4-expert" rules have similar but higher percentages of introduced errors, however, the error correction is lower than the two-expert rule (46.7\% for the three-expert rule and 33.35\% for the four-expert rule). Therefore, the higher the number of experts taken into account for belief propagation, the more resilient the group is to change the beliefs of its members, as for an autonomous systems to change its belief, all n-experts should have consensus and that might not happen so often for a high number of experts as seen in the three-expert and four-expert rules.

The Majority rule, as it can be observed, lead to very high percentage of introduced errors (50\%\%), and the lowest percentage of corrected errors (8\%), and this is due to the rules' nature, as the output of the majority between autonomous systems heavily depends on the quality attributes of the entire group forming the set of autonomous systems. The higher is the quality of attributes of autonomous systems in the group and the better the output of the majority rule, otherwise the rule can lead to poor results in terms of errors correction when using the collaboration. A detailed discussion about how the quality of the group influences collaborative reliability is presented in the next section. 


The Gravity-Point rule is similar to the Majority rule in terms of percentage of individual errors (25.6\%) as it applies the same concept as the majority rule, yet the Gravity-Point rule takes place between the closest (based on the quality of attributes) autonomous system  to the autonomous system where aggregation is performed, and that maintains the level of quality for the autonomous systems majority resulting in a lower percentage of introduced errors (30\%) than the Majority rule. However, for the less trustworthy autonomous system (with lower quality of attributes), the output of the sub-group majority is also determined by the quality of the sub-group, and that may result in degraded results during beliefs propagation.

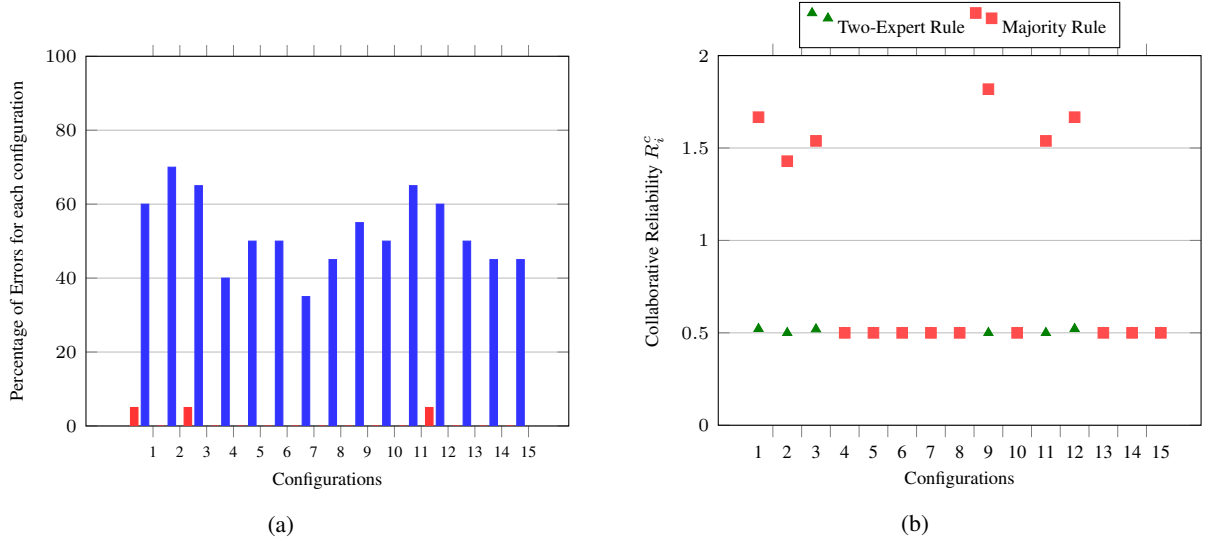
\begin{figure}
    \centering
    \begin{multicols}{2}
    \begin{subfigure}{0.45\textwidth}
    \vspace{15 pt}
    \begin{tikzpicture}
    \begin{axis}
    [
    ybar=0.4cm,
    bar width=0.1cm,
    width=8cm,
    height=6.5cm,
    ymin=0,
    ymax=100,
    yticklabel style={font=\scriptsize},
    xtick=data,
    xticklabels={1,2,3,4,5,6,7,8,9,10,11,12,13,14,15},
    xticklabel style={yshift=0ex, font=\tiny},
    x label style={at={(axis description cs:0.5,-0.1)},anchor=north, font=\scriptsize},
    ylabel={Percentage of Errors for each configuration},
    y label style={font=\scriptsize},
    xlabel={Configurations},
    ymajorgrids = true,
]

\addplot [fill=red!80, draw=red!80] coordinates {

    (1,5) 
    (2,0) 
    (3,5) 
    (4,0)
    (5,0)
    (6,0)
    (7,0)
    (8,0)
    (9,0)
    (10,0)
    (11,0)
    (12,5)
    (13,0)
    (14,0)
    (15,0)
};

\addplot [fill=blue!80, draw=blue!80] coordinates {

    (0,60) 
    (1,70) 
    (2,65) 
    (3,40)
    (4,50)
    (5,50)
    (6,35)
    (7,45)
    (8,55)
    (9,50)
    (10,65)
    (11,60)
    (12,50)
    (13,45)
    (14,45)
};


\end{axis}
    \end{tikzpicture}
    \centering
    \caption{} 
    \label{fig:percentage_leslie1}
    \end{subfigure}%
    
    \columnbreak

    \begin{subfigure}{0.45\textwidth}
    \begin{tikzpicture}
    \begin{axis}
    [ybar=0.4cm,
    bar width=0.2cm,
    width=8cm,
    height=6.5cm,
    ymin=0,
    ymax=2,
    yticklabel style={font=\scriptsize},
    xtick=data,
    xticklabels={1,2,3,4,5,6,7,8,9,10,11,12,13,14,15},
    xticklabel style={yshift=0ex, font=\tiny},
    x label style={at={(axis description cs:0.5,-0.1)},anchor=north, font=\scriptsize},
    ylabel={Collaborative Reliability $R_i^c$},
    y label style={font=\scriptsize},
    xlabel={Configurations},
    ymajorgrids= true,
    major x tick style={opacity=0},
    minor x tick num=1,
    minor tick length=1 ex,
    xticklabel style={yshift=0ex, font=\scriptsize},
    legend style={
    at={(0.50,1.02)},
    anchor=south,
    column sep=0ex,
    legend columns=-1,
    }
]
\addplot[only marks,mark=triangle*,color=black!50!green] plot coordinates{ 
    (1,0.5217391304) 
    (2,0.5) 
    (3,0.52) 
    (4,0.5)
    (5,0.5)
    (6,0.5)
    (7,0.5)
    (8,0.5)
    (9,0.5)
    (10,0.5)
    (11,0.5)
    (12,0.5217391304)
    (13,0.5)
    (14,0.5)
    (15,0.5)};
\addlegendentry{\scriptsize{Two-Expert Rule}};

\addplot[only marks,mark=square*,color=white!30!red!100] plot coordinates{ 
    (1,1.666666667) 
    (2,1.428571429) 
    (3,1.538461538) 
    (4,0.5)
    (5,0.5)
    (6,0.5)
    (7,0.5)
    (8,0.5)
    (9,1.818181818)
    (10,0.5)
    (11,1.538461538)
    (12,1.666666667)
    (13,0.5)
    (14,0.5)
    (15,0.5)};
\addlegendentry{\scriptsize{Majority Rule}};

\end{axis}
\end{tikzpicture}
\centering
\captionsetup{width=0.9\textwidth}
\caption{} 
\label{fig:crrleslie}
\end{subfigure}%
\end{multicols}
    
\caption{Detailed results per configuration. a) Each configuration has a different percentage of errors (blue bars), all percentages are more than 30\%, the percentage of errors from experts are also displayed (red bars). b) Results of collaborative reliability for each configuration for the two-expert and majority rule.}
\label{fig:leslie_percentage}
    
\end{figure}

We further detail the results of Fig.~\ref{fig:results} 
by selecting $15$ configurations shown in Fig. \ref{fig:leslie_percentage}, each configuration has different percentage of errors on beliefs that do not match the ground truth. Percentage of error that the two most experts are performing is also displayed in blue in Fig. \ref{fig:leslie_percentage}-(\subref{fig:percentage_leslie1}). We apply the two-expert and the majority rule for beliefs aggregation and propagation on each configuration. We quantify results using the previously defined metric \emph{Collaborative Reliability} (in Eq.,~\ref{eq:collab-reliab}). We observe that despite the high error rate, when applying the two-expert rule collaborative reliability is almost in all cases around the optimal value of $R_i^c = 0.5$ when (almost) all errors are corrected. This is due to the fact that despite the high error rate (max of 70\% for config number 2) correct beliefs from experts propagate and help correct erroneous beliefs in the group. This is not the case of the majority rule which allows to introduce errors leading to a collaborative reliability of $R_i^c \approx 1.5$ or higher for configurations where the percentage of errors is 50\% or higher.

\subsection{Effect of the 
Quality of the Group on Collaborative Reliability}
In a second series of experiments, we investigate 
the effect of the quality of attributes of the group on the number of errors introduced and errors corrected after applying beliefs aggregation and propagation rules with the collaboration.  

\begin{figure}
\centering
\includegraphics[width=0.6\textwidth]{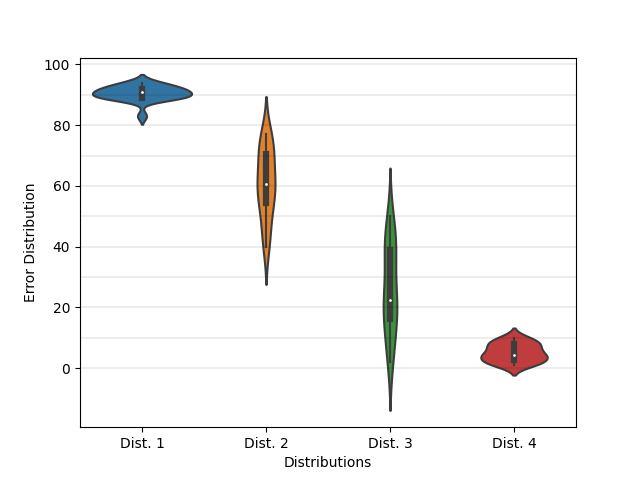}
\caption{Different Groups and their Error Distribution. Each group is generated independently by changing the quality of attributes of the autonomous systems in the group, different error distributions are obtained, as the lower the quality of attributes, the higher the error distribution is for each autonomous system.}
\label{fig:diff_groups}
\end{figure}

\subsubsection{Generation of Different Quality of Groups} As discussed previously, the quality of attributes affects the errors rate of an autonomous system when forming a belief (i.e., the higher is the quality of attributes, the lower is the error rate). We further generate multiple groups with different attributes quality, see Fig.~\ref{fig:diff_groups}, that results in different errors distributions, and each group is generated independently by varying the quality of attributes of its members. Dist.1 shows that the entire group has poor quality 
and Dist.4 is a group with all autonomous systems having a high quality of attributes.

\subsubsection{Effect on Collaborative Reliability}
\begin{figure}
\centering
\begin{multicols}{2}
\begin{subfigure}{0.35\textwidth}
\hspace{-20pt}
\begin{tikzpicture}
\begin{axis}[%
view={0}{90},
width=8cm,
height=7cm,
xmin=0, xmax=100,
xlabel={Configurations},
xlabel style = {font=\normalsize},
xticklabel style= {font=\scriptsize},
xticklabels={0,0, 20, 40, 60, 80, 100},
xtick distance=20,
ymin=0, ymax=1.6,
ylabel={Collaborative Reliability $R_i^c$}, label style={font=\scriptsize},
yticklabel style = {font=\scriptsize},
ymajorgrids,
legend style={at={(0.97,0.03)},anchor=south east,nodes=right}, font=\scriptsize]
\addplot[only marks, fill opacity=0, color=blue]  file{./Data/generated_numbers.tsv};

\addlegendentry{Two-Expert High-Quality};

\addplot[only marks, fill opacity=0, color=red] file{./Data/generated_numbers2.tsv};


\addlegendentry{Two-Expert Low-Quality};

\end{axis}
\end{tikzpicture}
\caption{}
\label{fig:col_reliability_2_exp_low_quality}
\end{subfigure}%

\columnbreak

\begin{subfigure}{0.45\textwidth}
\hspace{-13pt}
\vspace{-13pt}
\begin{tikzpicture}
\begin{axis}[%
view={0}{90},
width=8cm,
height=7cm,
xmin=1, xmax=8,
xlabel={},
xlabel style = {font=\scriptsize},
xtick=data,
xticklabels={Most-Expert, 2-Expert, 3-Expert, 4-Expert, Majority, Gravity-Point},
yticklabel style = {font=\scriptsize},
xticklabel style={yshift=0ex, rotate=90, font=\scriptsize},
x label style={anchor=north},
ymin=0, ymax=1.3,
ymajorgrids,
legend style={at={(0.97,0.03)},anchor=south east,nodes=right}, font=\scriptsize]
\addplot[only marks,mark=square*,color=blue] plot coordinates{ (2,1.037) (3,1.089) (4,1.101) (5,1.094) (6,1.12) (7,1.1015)};

\addlegendentry{Dist.1};

\addplot[only marks,mark=diamond*,color=yellow] plot coordinates{ (2,1.16853) (3,1.067) (4,1.1973) (5,1.0746) (6,1.1408) (7,1.1286)};

\addlegendentry{Dist.2};

\addplot[only marks,mark=triangle*,color=green] plot coordinates{ (2,0.6) (3,0.5) (4,0.62) (5,0.672) (6,0.63) (7,0.738)};

\addlegendentry{Dist.3};

\addplot[only marks,mark=square*,color=red] plot coordinates{ (2,0.54) (3,0.52) (4,0.692) (5,0.668) (6,0.557) (7,0.614) };

\addlegendentry{Dist.4};

\addplot [
color=black,
solid,
line width=1.0pt,
forget plot
]
coordinates{
 (1,1)(8,1) 
};
\end{axis}
\end{tikzpicture}%
\caption{}
\label{fig:col_reliability}
\end{subfigure}%
\end{multicols}
\vspace{-13pt}

\caption{(a) Collaborative Reliability by applying the Two-Expert Rule for the high-quality and low-quality group. The ratio for the low-quality group is always above $1$ resulting in a high Collaborative Reliability and the high-quality group has an average of $0.5$ Collaborative Reliability Ratio reflecting the low error distribution of the group. (b) Average Collaborative Reliability for Different Groups and Different Rules.
For each rule and each group, depending on the group's error distribution, the average collaborative reliability is calculated, and the lower the average collaborative reliability ratio, is the better.}
\label{fig:collaborative_rel_results}
\end{figure}
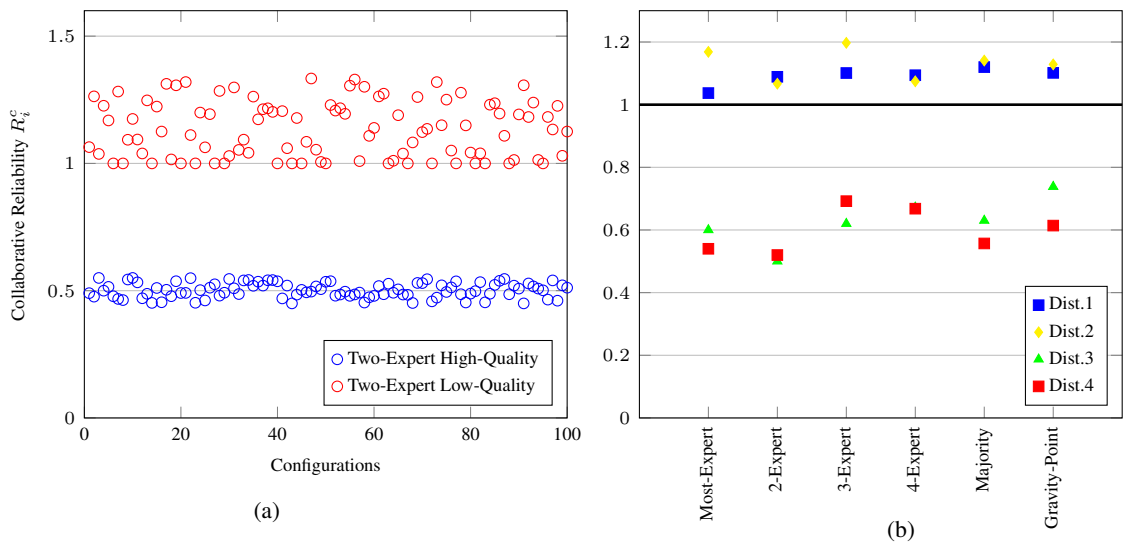

We investigate the effect of collaboration through beliefs aggregation and propagation using the previously defined rules over all four different groups with different distributions of quality of attributes. The goal is to increase reliability collaboratively and to increase the number of corrected errors while decreasing the number of introduced errors. We quantify results using the previously defined metric \emph{Collaborative Reliability} (in Eq.,~\ref{eq:collab-reliab}). For further clarification, the effect of the quality of the groups on the collaborative reliability ratio is shown in Fig.~\ref{fig:collaborative_rel_results}-(\subref{fig:col_reliability_2_exp_low_quality}), where the ratio is output to two different groups, a high-quality one and a low-quality one. Results are shown for all 100 configurations.

Note that according to the defined collaborative reliability metric, when the number of introduced errors and corrected errors is the same then the collaborative reliability ratio is equal to $R_i^c = 1$. The goal is as described previously: increase the number of corrected errors while minimizing the number of introduced errors. When all errors are corrected and no error is introduced, then the collaborative reliability ratio is optimally equal to $R_i^c = 0.5$. This is the case for the group of autonomous system with a high quality of attributes distribution (see Fig.~\ref{fig:collaborative_rel_results}-(\subref{fig:col_reliability_2_exp_low_quality})). Inversely, when the number of introduced errors is higher than the number of corrected errors then $R_i^c \geq 1$ as it is the case in the group of autonomous systems with low quality of attributes. Results are further summarized for the four groups with different qualities are summarized in Fig.~\ref{fig:collaborative_rel_results}-(\subref{fig:col_reliability}) where comparison between the groups is performed after applying all the rules.

It can be observed in Fig.~\ref{fig:collaborative_rel_results}-(\subref{fig:col_reliability}), 
for Distribution 1, the error distribution is the highest (ranging from 80\% to 95\%) resulting in an average collaborative reliability that is always above 1, along with Distribution 2, which has below-average quality of attributes (having an error distribution of 26\%-90\%). For these distributions, the average introduced errors is more than the average corrected ones resulting in the shown average Collaborative Reliability metric for all the rules in Fig.~\ref{fig:collaborative_rel_results}-(\subref{fig:col_reliability}). For Distribution 3, the groups' error distribution is evenly distributed as the group had autonomous systems making a low number of errors, and also autonomous systems making an error of almost 64\%. This error distribution successfully develop a system where an even hierarchy can be obtained, which is what we observe in any system in our environment, and Distribution 3 greatly resembles a real-life autonomous system where there is an acceptable range of errors that can be made by autonomous systems. In this distribution, it can be observed that the performance is best-of-all for the Two-Expert rule having an average Collaborative Reliability of almost $R_i^c \approx 0.5$, which means that almost all the individual errors before belief propagation are changed to corrected errors after belief propagation resulting in a $R_i^c = 0.5$ collaborative reliability. It can also be observed that the Most-Expert and Three-Expert rules have a much better performance in this group than the others meaning that the quality of the group greatly affects the resulting collaborative reliability. For Distribution 4, the groups' quality of attributes are the best resulting in a range of errors 0-22\%. The results of this group are greatly similar to the results of the hierarchy group (3), where the differences in the average ratios are very low as seen in Fig.~\ref{fig:collaborative_rel_results}-(\subref{fig:col_reliability}). However, the performance of the Two-Expert rule in the hierarchy group (3) still outperformed all the other rules and groups, which is greatly advantageous as this group is the most resembling a realistic diverse group. Also, it can be observed that for the high-quality group (4), the Majority rule and the gravity point rule outperform the Two-Expert rule as the error distribution is much lower for this group, and majority rule results in better output propagation than the hierarchy group (3).

\section{Discussion of Related Work}
\label{sec:rw}

We discuss in the following work and existing results, relevant to our contribution. We focus in particular on the following, 
i)The claim of availability of (compute and communication) infrastructure enabling to support autonomy and exchange of data between distributed autonomous systems, ii) The  issue of peer disagreement in the field of collective intelligence and main results on aggregation rules based on consensus and majority, iii) The term collaborative autonomy already established in the literature and that recognizes different capabilities to autonomous systems, iv)The term trustworthiness and how it is used in the literature and the link to reliability, v) Main results from the  work of leslie lamport in his seminal paper "The Byzantine Generals Problem" on reaching agreement in classical distributed systems in the presence of conflicting exchanged information resulting from faulty processors.

\paragraph{Infrastructure for Cooperative Autonomous Systems}
Given quickly increasing connectivity and available computation capabilities (e.g., using advanced 5G/6G wireless communication technology and Edge-Cloud-Infrastructures), more future autonomous systems will depend on cooperation in combination with infrastructure~\cite{liu2022ieee}. 
Cooperative driving currently leverages Vehicle-to-X (i.e., vehicle to vehicle and/or vehicle to
infrastructure) communication technologies aiming to carry out cooperative functionalities, such
as cooperative sensing and cooperative maneuvering. 
Infrastructure to support Device-Edge-Cloud Continuum enables sharing of information between distributed (autonomous)
systems and infrastructure~\cite{MundhenkHHZ22,keynote_Date2023}. Broadcasting information from every vehicle using for instance collective awareness messages is thereby enabled. Every vehicle can later make a decision locally or globally, like maneuvering, based on this information. 
The fact, however, that there is an infrastructure to share information does by no means imply that the issue of safety and correct collective behavior is assured.
There is a lack of structured and systematic processes and methods to collectively reason about shared information in order to improve safety, trustworthiness and good
decision making using collaboration.

\paragraph{Collective Intelligence}
Collective intelligence in social science refers to the collective exploitation of distributed intelligence for a better decision making in groups as opposed to individual decision making. Collective intelligence as defined in~\cite{collective_intelligence_def} is a form of universally distributed intelligence, constantly enhanced, coordinated in real-time, and resulting in the effective mobilization of skills. Collective intelligence therefore builds on the premise that a group is collectively "smarter" than a single individual, which globally leads to better decision making processes. Collective intelligence has been exploited in different engineering domains~\cite{collective_intelligence} with a particular focus in social computing is on human-machine interaction where interconnected groups of people and computers are required to collectively perform intelligent actions. 
Social epistemology, and in particular belief propagation, deals with the study of how "good" knowledge and belief propagate in a society~\cite{sep_epistemology_social}. 
We consider in this paper similar concepts where we exploit collaboration between multiple autonomous systems in order to design a good collective decision making process. 
%
In the past decades, social choice theory and judgement aggregation have been extensively studied in various domains such as philosophy, welfare economics, AI, and multi-agent systems, in order to provide a principled definition of the aggregation of individual attitudes into a social or collective attitude~\cite{logic_collective_Reasoning}, ~\cite{impossibility_theorem}. One of the most fundamental results in the field of judgement aggregation theory in social science is the \emph{impossibility theorem}, that states there is no aggregation procedure that can guarantee both rationality of the outcome and the fairness of the aggregation at the same time ~\cite{impossibility_theorem}. The proof of this result is mainly based on the majority voting exploiting Arrow’s impossibility theorem, or Arrow’s paradox~\cite{sep-arrows-theorem}. 
Even if the goal of this paper is not, yet, to prove properties of aggregation rules, it is worth to mention that consensus and majority aggregation rules are dominant in classical collective reasoning theory. The  majority aggregation rule is actually the main assumptions to seminal results around the impossibility theorem. However, even if the majority rule offers some notion of fairness considering no differentiation between contributing individuals, and can work good in practice, we have seen in this paper that it does not always lead to high collaborative reliability results (e.g., compared to the two-expert based rule) and will necessarily come at a high communication overhead at operation time using infrastructure to share data.

\paragraph{Collaborative Autonomy}
The idea of looking at autonomous systems as having different capabilities has been already referred to in the context of collaborative systems. The term collaborative autonomy as defined by Zann Gill in~\cite{Gill11} as the principle underpinning collaborative intelligence through which individual contributors maintain their roles and priorities, as they apply their unique skills and leadership autonomy in a problem-solving process.  
Collaborative autonomy therefore recognizes that autonomous systems have different capabilities, of for instance, pattern recognition, and a consensus in reaching collaborative intelligence (as traditionally applied in autonomous multi-agent systems)~\cite{multi-agent} is not always required. Collaborative autonomy thereby explicitly acknowledges that systems have different quality attributes when they collectively cooperate, and this is the core concept on which we base our contribution in this paper. This distinction may seem minor but is fundamental when aggregating knowledge and actions of autonomous systems using collective intelligence. We showed in this paper that to provide an efficient and more trustworthy decision making processes, it is possible to exploit individual attributes (e.g., sensors quality) of autonomous systems and formulates based on that rules for collaborative autonomous systems. 

\paragraph{Trustworthiness in Autonomous Systems}
When referring to autonomous systems, the term trustworthiness can be defined considering different parameters and there is  not a unified definition in the literature of what trustworthiness actually means. A widely accepted definition of trust is lacking in both social science literature and human-machine interaction literature~\cite{adams2001trust}. 
The author in~\cite{devitt2018trustworthiness} argues that trustworthiness can refer to an intrinsic property of an agent (similar to height or a relation property of tallness) which is equivalent to what we consider in this paper as attributes and their quality intrinsic to an autonomous system. The work establishes as well a link between trustworthiness and reliability where a system is considered to be trustworthy if "we can rely on it being right".  An unreliable sensing or perception system (i.e., with a high number of perception errors) can be considered as untrustworthy. This might be (from an engineering perspective) one of the simplest and accurate way to define trustworthiness and this is what we as well consider in this paper. 
%
The notion of trustworthiness has been as well used in the context of social assessments where trustworthiness is understood as a dispositional and relational property that can be established by combining judgments from multiple
agents, such as through peer assessment~\cite{trust_automation}. This notion follows how we use trustworthiness in this paper to evaluate, through beliefs aggregation and propagation, whether a group of autonomous systems is providing correct statements on the environment or not. 
Another related topic in autonomous systems is calibrated trust. Calibrated trust is the measure of how appropriate it is for the operator or user to trust a given system~\cite{kok2020trust}. 
Calibrated trust is about finding the right balance between over- and under-confidence to get optimal results~\cite{mcdermott2019practical}~\cite{10.1145/2667317.2667330}. 
As most automation does not have a constant performance, situations where the system excels or underperforms are used as calibration points. 
The main issue in the literature is often to provide evidence for trustworthiness. In~\cite{mcdermott2019practical}, trust is determined by a human equivalent behavior of the system (i.e, the system behaves as a human would behave) in certain situations. By using in this paper quality attributes, we provide a \emph{systematic} way to provide evidences for trustworthiness. In the case of quality attributes changing over time (e.g., perception angle or distance), this can as well be used as a good measure for calibrating trust at specific defined operation points. 




\paragraph{Reliability in Classical Distributed Systems}
%
An important fundamental result from distributed computing is attributed to Leslie Lamport in his formulation of his famous Byzantine Generals Problem~\cite{lamport2019byzantine}, where a set of distributed computer systems communicating through messages must cope with the (malicious or not) failure of one or more of its components, where conflicting information are being sent to other systems. In this case, it becomes exceedingly difficult to distinguish faulty statements from non-faulty ones, based on collectively gathered information. In his paper, "Reaching Agreement in the Presence of Faults"~\cite{leslie_lamport_errors}, Leslie Lamport further shows that if the number of faulty statements is third of the number of total statements (e.g., three computer systems and only one is malicious) that it will always be possible to infer correct information (i.e., what is equivalent to the ground truth). No agreement can be found in the presence of at least $n = 3m+1$ or greater of systems if $m$ of them are faulty and  in the presence of an unknown subset of faulty processors which are sending incorrect information, the problem becomes unsolvable. 
Interestingly enough, existing results often consider (implicitly or explicitly) homogeneous systems like in classical distributed computing and in multi-agent autonomous systems and do not exploit sufficiently the differences between systems that can affect the quality of collected information when aggregated.
The main assumptions to the Byzantine Generals Problem are also homogeneity of systems and consensus as aggregation rule.  One key aspect to consider is that solving this problem can be expensive in terms of both time and the number of messages. When assessing this against our work defining aggregation rules based on autonomous systems that are more trustworthy (i.e., experts), our approach is characterized by its requirement of relatively few messages (propagation of beliefs from more experts to less experts) and is not as computationally expensive considering the defined applied reduction rules of the binary decision trees for beliefs aggregation and propagation.
By relying on fewer experts in the group as we propose, we can also see that we can achieve high collaborative reliability results as shown in the evaluation even in the presence of a large number (more than third) of autonomous systems making an error.  

\section{Concluding Remarks and Key Takeaways}
\label{sec:concl}
We presented in this paper a general approach to increase good decision making in autonomous systems using collaboration through data sharing. We exploit in particular quality attributes of autonomous systems as a measure and evidence of trustworthiness to decide which autonomous systems are more trustworthy than others. Trustworthy autonomous systems are referred to as experts and are used to define rules for beliefs aggregations and propagation to correct other autonomous systems erroneous beliefs in the perception of the environment. 
We further exploit ranking of autonomous systems, formally modelled as lattices, using quality attributes to define an order of beliefs aggregation and propagation in defined BDDs, and apply reduction rules for simplifying ordered BDDs where, depending on the aggregation rule, convergence of beliefs of autonomous systems of all (or the rest of the group) can be achieved. We evaluate our results considering multiple generated benchmarks and configurations where different rules are applied. We compare results using our defined metric: collaborative reliability, as a measure to assess the ratio between introduced and corrected errors after the collaboration. 
We summarize in the following important observations and key outcomes. 

\begin{enumerate}
    \item As shown in the evaluation section, the two-experts rule offers the best compromise between a high collaborative reliability ratio and a high efficiency since reduction rules lead to a more simplified form of binary decision diagrams as proven by the convergence of beliefs. In this case, the complexity of the binary decision diagrams is reduced from $2^{n+1}$ for a group of $n$ autonomous systems to $O(n)$. 
    In fact, no matter how large is the group of autonomous systems, by applying the two-expert rule recursively, it becomes enough to identify the two most experts in the group and if both have similar beliefs then this latter is adopted by the rest of the group, which leads to a more efficient decision making process compared to consensus or the majority rule. 
   
    \item The two-expert rule leads to better reliability considering the collaboration than the majority rule under the assumption we consider in this paper that experts or more trustworthy autonomous systems have better quality of attributes and are therefore less likely to make errors than less trustworthy ones.  Increasing the number of experts in the applied rule only allows to increase resiliency (i.e., likelihood that all experts are making errors simultaneously.)  
    It becomes relevant to consider the benefit of using more than two experts for the aggregation rule by comparing the level of resiliency between two experts or more. 

    \item We have observed experimentally that relying on fewer trustworthy autonomous systems (instead of the entire group) allows to improve reliability using the collaboration even in the presence of a high percentage of erroneous beliefs. 
    Even if we do not have yet a formally proven result regarding whether better guarantees can be formulated in the presence of higher error percentages (e.g., as compared to the results of Leslie Lamport in the presence of more than 30\% of erroneous beliefs), experiments show a potential of this approach in dealing with conflicting information to reach a consensus as through convergence of beliefs.

    \item Another interesting observation is that the quality of the group actually matters when using the collaboration. 
   This is somehow intuitive since when the quality of the entire group is poor, collaboration is not helping much in improving collectively decision making. It becomes as well apparent that relying on most experts in the group brings a benefit when there is a high discrepancy in the distribution of quality of attributes of the group. If not, then reliability as demonstrated in this paper is not significantly improved by using collaboration. 

\end{enumerate}

\paragraph{On the practicability of the proposed approach} in this paper, we proposed efficient formal models like lattices and reduced binary decision diagrams that we believe have great potential to ease computability and allow a fast decision processing at operation time. One key factor to the success of this approach in practice is the choice of appropriate attributes that are considered to determine trustworthiness and ranking of autonomous systems. Mapping of for instance physical characteristics of different sensors types and how detection performs based on distance to the object (e.g., Lidar versus optic) to characterize errors and percentage on quality of attributes is crucial. Moreover, we considered throughout the paper only two examples of attributes, namely perception angle and distance.  In practice, many more attributes can be considered leading to a much higher complexity of the lattice. Identifying few number of key attributes and their quality that actually matter will be key in keeping the dimensionality of the lattice low as well as the size of sub-group of trustworthy autonomous systems on which to rely for decision making. We believe this will heavily depend on the application domain and the context of operation.  
\bibliographystyle{plain}
\bibliography{references}

\begin{thebibliography}{10}

\bibitem{adams2001trust}
Barbara~Dale Adams, Robert~D Webb, and David~J Bryant.
\newblock {\em Trust in teams: Literature review}.
\newblock Humansystems Incorporated, 2001.

\bibitem{bdd-seminal}
Randal~E. Bryant.
\newblock Symbolic boolean manipulation with ordered binary-decision diagrams.
\newblock {\em ACM Comput. Surv.}, 24(3):293–318, sep 1992.

\bibitem{multiagent}
Louise~A. Dennis and Michael Fisher.
\newblock Verifiable self-aware agent-based autonomous systems.
\newblock {\em Proceedings of the IEEE}, 108(7):1011--1026, 2020.

\bibitem{devitt2018trustworthiness}
S~Devitt.
\newblock Trustworthiness of autonomous systems.
\newblock {\em Foundations of trusted autonomy (Studies in Systems, Decision and Control, Volume 117)}, pages 161--184, 2018.

\bibitem{keynote_Date2023}
Dirk Elias, Dirk Ziegenbein, Philipp Mundhenk, Arne Hamann, and Anthony Rowe.
\newblock The cyber-physical metaverse - where digital twins and humans come together.
\newblock In Ian O'Connor, Robert Wille, and Ioana Vatajelu, editors, {\em 2023 Design, Automation {\&} Test in Europe Conference {\&} Exhibition, {DATE} 2023, Antwerp, Belgium, April 17-19, 2023}. {IEEE}, 2023.

\bibitem{Gill11}
Zann Gill.
\newblock Collaborative intelligence in living systems: algorithmic implications of evo-devo debates.
\newblock In Natalio Krasnogor and Pier~Luca Lanzi, editors, {\em 13th Annual Genetic and Evolutionary Computation Conference, {GECCO} 2011, Companion Material Proceedings, Dublin, Ireland, July 12-16, 2011}, pages 803--804. {ACM}, 2011.

\bibitem{sep_epistemology_social}
Alvin Goldman and Cailin O’Connor.
\newblock {Social Epistemology}.
\newblock In Edward~N. Zalta, editor, {\em The {Stanford} Encyclopedia of Philosophy}. Metaphysics Research Lab, Stanford University, {W}inter 2021 edition, 2021.

\bibitem{hecht2023optik}
Eugene Hecht.
\newblock {\em Optik}.
\newblock Walter de Gruyter GmbH \& Co KG, 2023.

\bibitem{kok2020trust}
Bing~Cai Kok and Harold Soh.
\newblock Trust in robots: Challenges and opportunities.
\newblock {\em Current Robotics Reports}, 1:297--309, 2020.

\bibitem{lamport2019byzantine}
Leslie Lamport, Robert~E. Shostak, and Marshall~C. Pease.
\newblock The byzantine generals problem.
\newblock In Dahlia Malkhi, editor, {\em Concurrency: the Works of Leslie Lamport}, pages 203--226. {ACM}, 2019.

\bibitem{trust_automation}
John~D. Lee and Katrina~A. See.
\newblock Trust in automation: Designing for appropriate reliance.
\newblock {\em Human Factors}, 46(1):50--80, 2004.
\newblock PMID: 15151155.

\bibitem{collective_intelligence_def}
Pierre Levy and Robert Bononno.
\newblock {\em Collective Intelligence: Mankind's Emerging World in Cyberspace}.
\newblock Perseus Books, USA, 1997.

\bibitem{impossibility_theorem}
Christian List and Philip Pettit.
\newblock Aggregating sets of judgments: An impossibility result.
\newblock {\em Economics \&amp; Philosophy}, 18(1):89–110, 2002.

\bibitem{liu2022ieee}
Shaoshan Liu and Jean-Luc Gaudiot.
\newblock Ieee international roadmap for devices and systems (irds) autonomous machine computing white paper, 2022.

\bibitem{collective_intelligence}
Thomas~W. Malone and Michael~S. Bernstein.
\newblock {\em Handbook of Collective Intelligence}.
\newblock The MIT Press, 2015.

\bibitem{mcdermott2019practical}
Patricia~L McDermott and Ronna N~ten Brink.
\newblock Practical guidance for evaluating calibrated trust.
\newblock In {\em Proceedings of the Human Factors and Ergonomics Society Annual Meeting}, volume~63, pages 362--366. SAGE Publications Sage CA: Los Angeles, CA, 2019.

\bibitem{sep-arrows-theorem}
Michael Morreau.
\newblock {Arrow’s Theorem}.
\newblock In Edward~N. Zalta, editor, {\em The {Stanford} Encyclopedia of Philosophy}. Metaphysics Research Lab, Stanford University, {W}inter 2019 edition, 2019.

\bibitem{MundhenkHHZ22}
Philipp Mundhenk, Arne Hamann, Andreas Heyl, and Dirk Ziegenbein.
\newblock Reliable distributed systems.
\newblock In Cristiana Bolchini, Ingrid Verbauwhede, and Ioana Vatajelu, editors, {\em 2022 Design, Automation {\&} Test in Europe Conference {\&} Exhibition, {DATE} 2022, Antwerp, Belgium, March 14-23, 2022}, pages 287--291. {IEEE}, 2022.

\bibitem{leslie_lamport_errors}
M.~Pease, R.~Shostak, and L.~Lamport.
\newblock Reaching agreement in the presence of faults.
\newblock {\em J. ACM}, 27(2):228–234, apr 1980.

\bibitem{logic_collective_Reasoning}
Daniele Porello.
\newblock Logics for collective reasoning.
\newblock In Andreas Herzig and Emiliano Lorini, editors, {\em Proceedings of the European Conference on Social Intelligence (ECSI-2014), Barcelona, Spain, November 3-5, 2014}, volume 1283 of {\em {CEUR} Workshop Proceedings}, pages 148--159. CEUR-WS.org, 2014.

\bibitem{10.1145/2667317.2667330}
Christina R\"{o}del, Susanne Stadler, Alexander Meschtscherjakov, and Manfred Tscheligi.
\newblock Towards autonomous cars: The effect of autonomy levels on acceptance and user experience.
\newblock In {\em Proceedings of the 6th International Conference on Automotive User Interfaces and Interactive Vehicular Applications}, AutomotiveUI '14, page 1–8, New York, NY, USA, 2014. Association for Computing Machinery.

\bibitem{multi-agent}
Michael Wooldridge.
\newblock {\em An Introduction to Multiagent Systems}.
\newblock Wiley, Chichester, UK, 2 edition, 2009.

\end{thebibliography}

\end{document}